\documentclass[preprint]{uai2022}

\usepackage[american]{babel}

\usepackage{natbib} 
\usepackage{mathtools} 
\usepackage{booktabs} 
\usepackage{tikz} 

\usepackage{amsfonts,amssymb}
\usepackage{graphicx,xcolor}
\usepackage{wrapfig}
\newtheorem{theorem}{Theorem}
\newtheorem{lemma}{Lemma}
\newtheorem{definition}{Definition}
\newcommand{\KL}{{\rm KL}}     
\newcommand{\CE}{\mathcal{CE}} 
\renewcommand{\H}{\mathcal{H}} 
\newcommand{\N}{\mathcal{N}}   
\newcommand{\MI}{\mathcal{I}}  
\newcommand{\E}{\mathbb{E}}    
\renewcommand{\L}{\mathcal{L}} 
\newcommand{\q}{{\rm q}}       
\newcommand{\p}{{\rm p}}       
\newcommand{\Z}{{\rm Z}}       
\newcommand{\x}{\mathbf{x}}    
\newcommand{\randomt}{\mathbf{t}}    
\newcommand{\rtheta}{{\color{red}\theta}}
\newcommand{\bmu}{\boldsymbol{\mu}}
\newcommand{\bsigma}{\boldsymbol{\sigma}}
\newcommand{\D}{\mathcal{D}}   
\newcommand{\m}{{\rm m}}       
\newcommand{\rd}{{\rm d}}      
\newcommand{\FIM}{\mathcal{F}} 
\newcommand{\thetastar}{\theta^*} 
\newcommand{\Sigmastar}{{\Sigma^*}} 
\DeclareMathOperator*{\argmin}{arg\,min}
\newcommand{\qed}{\hfill\ensuremath{\blacksquare}}

\title{Interpolating Between Sampling and Variational Inference with Infinite Stochastic Mixtures}

\author[1]{\href{mailto:lange.richard.d@gmail.com}{Richard D. Lange}{}}
\author[1]{Ari S. Benjamin}
\author[2]{Ralf M. Haefner$^*$}
\author[3]{\href{mailto:xaq@rice.edu}{Xaq Pitkow$^*$}{}}
\affil[1]{%
    Dept. of Neurobiology\\
    University of Pennsylvania\\
    Philadelphia, Pennsylvania, USA
}
\affil[2]{%
    Dept. of Brain and Cognitive Sciences\\
    University of Rochester\\
    Rochester, New York, USA
}
\affil[3]{%
    Baylor College of Medicine\\
    Rice University
    Houston, Texas, USA
  }

\begin{document}

\maketitle

\begin{abstract}
Sampling and Variational Inference (VI) are two large families of methods for approximate inference that have complementary strengths. Sampling methods excel at approximating arbitrary probability distributions, but can be inefficient. VI methods are efficient, but may misrepresent the true distribution. Here, we develop a general framework where approximations are stochastic mixtures of simple component distributions. Both sampling and VI can be seen as special cases: in sampling, each mixture component is a delta-function and is chosen stochastically, while in standard VI a single component is chosen to minimize divergence. We derive a practical method that interpolates between sampling and VI by solving an optimization problem over a mixing distribution. Intermediate inference methods then arise by varying a single parameter. Our method provably improves on sampling (reducing variance) and on VI (reducing bias+variance despite increasing variance). We demonstrate our method's bias/variance trade-off in practice on reference problems, and we compare outcomes to commonly used sampling and VI methods. This work takes a step towards a highly flexible yet simple family of inference methods that combines the complementary strengths of sampling and VI.
\end{abstract}

\section{Introduction}

\begin{figure*}[ht]
    \centering
    \includegraphics[width=6.5in]{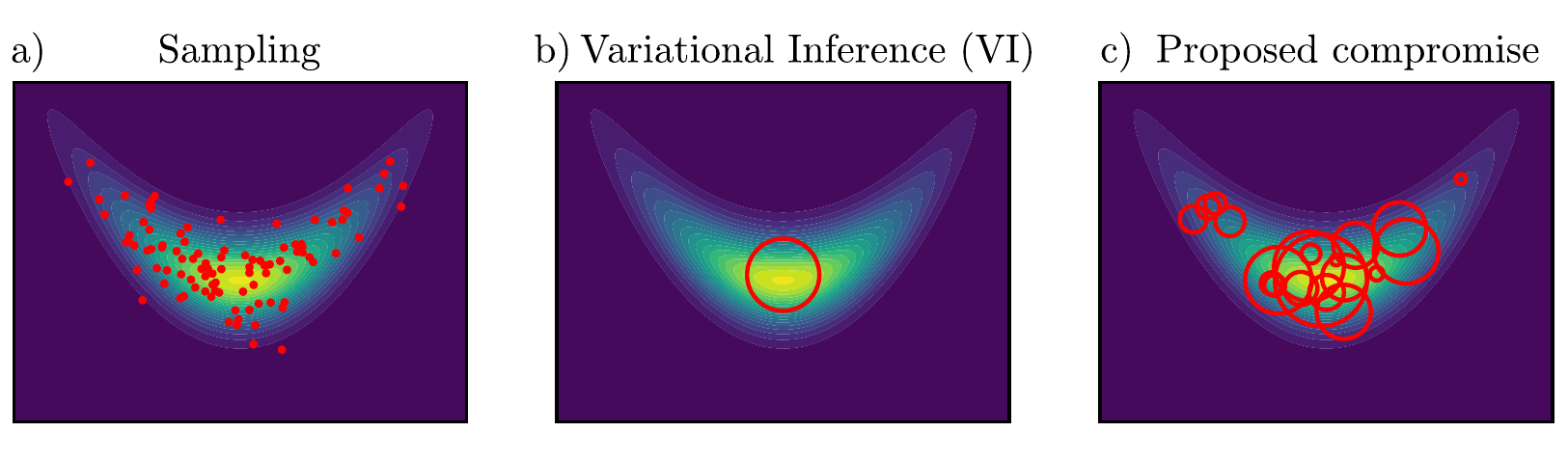}
    \caption{Conceptual introduction on a toy 2D example.
    \textbf{a)} Sampling methods approximate the underlying $\p(\x)$ with a stochastic set of representative points.
    \textbf{b)} Variational Inference (VI) methods begin by selecting an approximating distribution family, $\q(\x;\theta)$, here an isotropic Gaussian plotted as an ellipse at its $1\sigma$ contour. The optimal parameters $\thetastar$ are chosen to minimize $\KL(\q(\x;\theta)||\p(\x))$.
    \textbf{c)} We propose using a stochastic mixture of component distributions, where \emph{parameters} $\theta$ are sampled rather than the variable of interest $\x$.
    }
    \label{fig:concept}
\end{figure*}

We are concerned with the familiar and general case of approximating a probability distribution, such as occurs in Bayesian inference when both the prior over latent variables and the likelihood function connecting them to data are known, but computing the posterior exactly is intractable. There are two largely separate families of techniques for approximating such intractable inference problems: Markov Chain Monte Carlo (MCMC) sampling, and Variational Inference (VI) \citep{Bishop2006,Murphy2012}.

Sampling-based methods, including MCMC, approximate a distribution with a finite set of representative points. MCMC methods are stochastic and sequential, generating a sequence of sample points that, given enough time, become representative of the underlying distribution increasingly well. MCMC sampling is (typically) asymptotically unbiased, at the expense of high variance, leading to long run times in practice. Similar to the approach we take here, sampling methods are studied at different scales: both in terms of their asymptotic limit (i.e. their bias at infinitely many samples) and their practical behavior for finite samples or other resource limits \citep{Korattikara2013,Angelino2016}.

Variational Inference (VI) refers to methods that produce an approximate distribution by minimizing some quantification of divergence between the approximation and the desired posterior distribution \citep{Blei2017,Zhang2019}. For the purposes of this paper, we will use VI to refer to the most common flavor of variational methods, namely minimizing the Kullback-Leibler ($\KL$) divergence between an approximate distribution from a fixed family and the desired distribution \citep{Bishop2006,Wainwright2008,Murphy2012,Blei2017}. 
The best-fitting approximate distribution is often used directly as a proxy for the true posterior in subsequent calculations, which can greatly simplify those downstream calculations if the approximate distribution is itself easy to integrate. In contrast to MCMC, VI is often used in cases where speed is more important than asymptotic bias \citep{Angelino2016,Blei2017,Zhang2019}.

In this work, our goal is to develop an intermediate family of methods that ``interpolate'' between MCMC and VI, inspired by a simple and intuitive picture (Figure \ref{fig:concept}): we propose applying sampling methods \emph{in the space of variational parameters} such that the resulting approximation is a stochastic mixture of variational ``component'' distributions \citep{Yin2018}. This extends sampling by replacing the sampled points with extended components, 
and it extends VI by replacing the single best-fitting variational distribution with a stochastic mixture of more localized components. This is qualitatively distinct from previous variational methods that use \emph{stochastic optimization}: rather than stochastically optimizing a single variational approximation \citep{Hoffman2013,Salimans2015}, we use stochasticity to construct a \emph{random mixture} of variational components that achieves lower asymptotic bias than any one component could. As we will show below, this framework generalizes both sampling and VI, where sampling and VI emerge as special cases of a single optimization problem.

This paper is organized as follows. In section \ref{sec:setup}, we set up the problem and our notation, and describe how both classic sampling and classic VI can be understood as special cases of stochastic mixtures. In section \ref{sec:framework}, we introduce an intuitive framework for reasoning about infinite stochastic mixtures and define an optimization problem that captures the trade-off between sampling and VI. Section \ref{sec:algorithm} introduces an approximate objective and closed-form solution and describes a simple practical algorithm. Section \ref{sec:bias_variance} gives empirical and theoretical results that show how our method interpolates the bias and variance of sampling and VI. Finally, section \ref{sec:discussion} concludes with a summary, related work, limitations, and future directions.

\section{SETUP AND NOTATION}\label{sec:setup}

Let $\p^*(\x)=\Z\p(\x)$ denote the unnormalized probability distribution of interest, with unknown normalizing constant $\Z$. For instance, in the common case of a probabilistic model with latent variables $\x$, observed data $\D$, and joint distribution $\p(\x,\D)$, we are interested in approximations to the posterior distribution $\p(\x|\D)$. This is intractable in general, but we assume that we have access to the un-normalized posterior $\p^*(\x|\D) = \frac{1}{\Z} \p(\D|\x) \p(\x)$.\footnote{To reduce clutter, $\D$ will be dropped in the remainder of the paper, and we will use only $\p(\x)$ and $\p^*(\x)$.} 
Let $\q(\x;\theta)$ be any ``simple'' distribution that may be used used in a classic VI context (such as mean-field or Gaussian), and let $\m_T(\x)$ be a mixture containing $T$ of these simple distributions as components, defined by a set of $T$ parameters $\lbrace{\theta^{(1)}, \ldots, \theta^{(T)}\rbrace}$:
\begin{equation}\label{eqn:define_m_t}
    \m_T(\x) \equiv \frac{1}{T} \sum\limits_{t=1}^T \q(\x ; \theta^{(t)}) \, .
\end{equation}
For example, if $\q$ is a multivariate normal with mean $\mu$ and covariance $\Sigma$, then $\theta^{(t)}=\lbrace{\mu^{(t)},\Sigma^{(t)}}\rbrace$ and $\m_T(\x)$ would be a mixture of $T$ component normal distributions \citep{Gershman2012b}.

We will study properties of distributions over component parameters, which we denote $\psi(\theta)$ \citep{Ranganath2016}. If the set of $\theta^{(t)}$ is drawn randomly from $\psi(\theta)$, then as $T \rightarrow \infty$, $\m_T(\x)$ approaches the idealized infinite mixture,
\begin{equation}\label{eqn:define_m}
    \m(\x) \equiv \int_\theta \q(\x;\theta) \psi(\theta) \rd \theta \, .
\end{equation}


\paragraph{Sampling and VI as special cases of the mixing distribution.} Let $\thetastar=\argmin_\theta \KL(\q(\x;\theta)||\p(\x))$ be the parameters corresponding to the classic single-component variational solution. VI corresponds to the special case where the mixing distribution $\psi(\theta)$ is a Dirac delta around $\thetastar$, or  $\psi(\theta) = \delta(\theta-\thetastar)$, in which case the mixture $\m_T(\x)$ is equivalent to $\q(\x;\thetastar)$ regardless of the number of components $T$. Sampling can also be seen as a special case of $\psi(\theta)$ in which each component narrows to a Dirac delta ($\psi(\theta)$ places negligible mass on regions of $\theta$-space where components have appreciable width), and the means of the components are distributed according to $\p(\x)$. This requires that the component family $\q(\x;\theta)$ is capable of expressing a Dirac-delta at any point $\x$, such as a location-scale family. Thus, both sampling and VI can be seen as limiting cases of stochastic mixture distributions, $\m_T(\x)$, defined by a distribution over component parameters, $\psi(\theta)$. In what follows, we will show how designing the mixing distribution $\psi(\theta)$ allows us to create mixtures that trade-off the complementary strengths of sampling and VI. 

\begin{figure*}[ht]
    \centering
    \includegraphics[width=\textwidth]{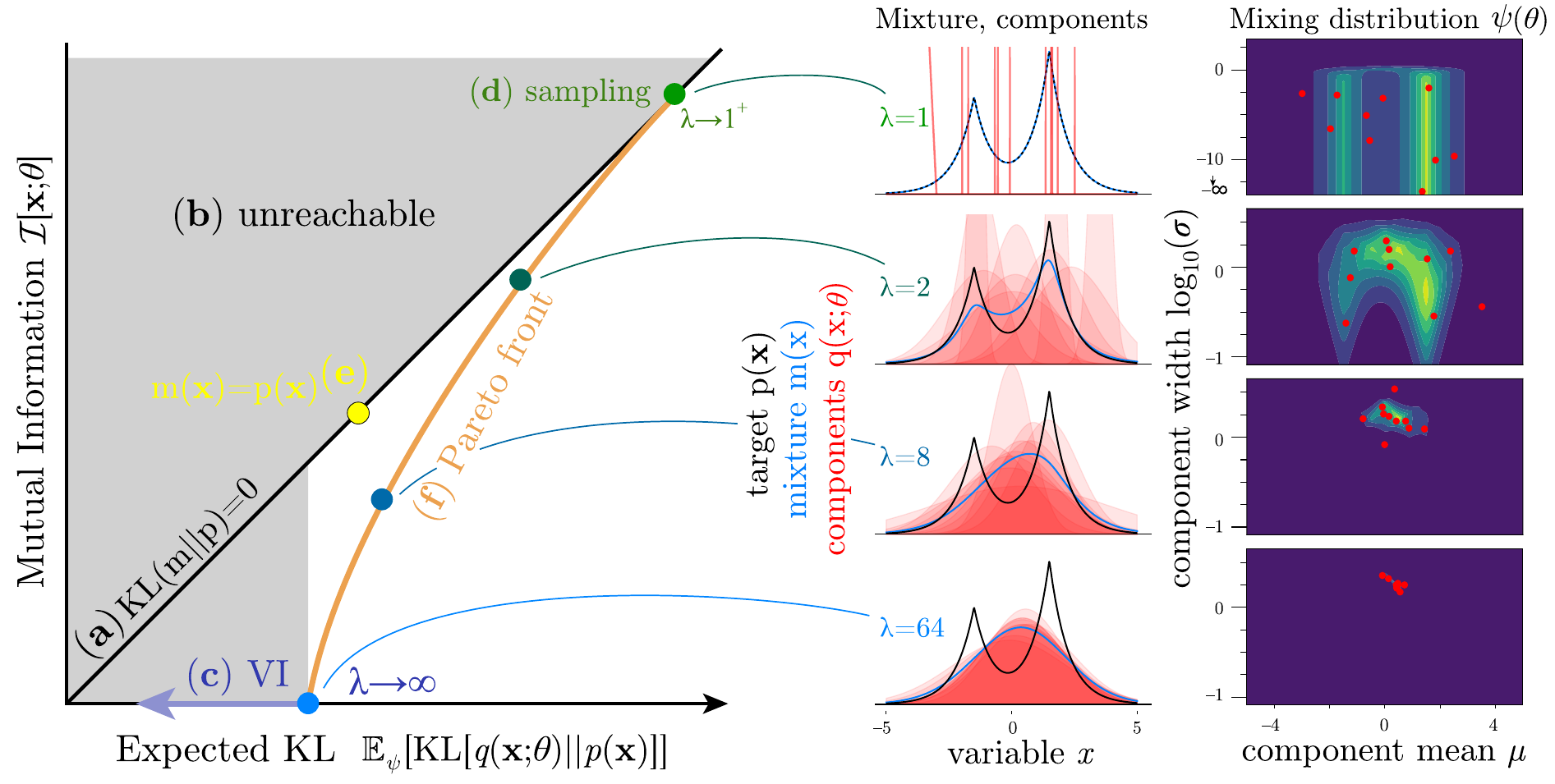}
    \caption{\textit{Left}: Understanding mixtures in terms of Mutual Information and Expected KL.
    \textbf{a)} The quality of any infinite mixture (in terms of $\KL(\m||\p)$) is given by its distance from the y=x line (black diagonal line).
    \textbf{b)} Two unreachable regions are shaded in gray: above the y=x line (because $\KL(\m||\p) \geq 0$), and to the left of the single-component variational solution, since VI achieves the minimum $\KL(\q||\p)$.
    \textbf{c)} When $\psi(\theta)=\delta(\theta-\thetastar)$ as in classic VI, Expected KL is at its minimum and Mutual Information is zero. 
    Increasing the expressiveness of $\q$ corresponds to moving left along the x-axis (blue arrow).
    \textbf{d)} Because sampling is unbiased, it is a mixture that lives on the $\KL(\m||\p)=0$ or $y=x$ line. If $\x$ is discrete, the coordinates of the point marked (d) are $(\H[\x], \H[\x])$, i.e. the entropy of $\p(\x)$. When $\x$ is continuous, both Mutual Information and Expected KL grow unboundedly together as the individual components narrow. 
    \textbf{e)} Any point on the y=x line implies $\m(\x)=\p(\x)$, and this may be possible without resorting to sampling for certain combinations of $\p$ and $\q$. However, such mixtures are not guaranteed to exist for all problems, and are difficult to find due to the intractability of Mutual Information.
    \textbf{f)} We propose a family of mixture approximations, parameterized by $\lambda$, that connects VI to sampling in a natural and principled way. Points on this curve correspond to solutions to the (approximate version of the) objective in (\ref{eqn:weighted_optim}).
    \textit{Middle}: Examples in a 1D toy problem, where $\p(\x)$ is an unequal mixture of two heavy-tailed distributions (black lines), and $\q(\x;\theta)$ is a single Gaussian component with parameters $\theta=\lbrace{\mu, \log\sigma}\rbrace$ (transluscent red components). \textit{Right}: Varying $\lambda$ controls the mixing distribution over $\theta$ (contours). Red points correspond to the Gaussian components in the middle.
    }
    \label{fig:mi_kl_space}
\end{figure*}

\section{Conceptual Framework}\label{sec:framework}

\subsection{Decomposing $\KL(\m||\p)$ Into Mutual Information and Expected KL}

The idealized infinite mixture $\m(\x)$ is fully defined by the chosen component family $\q(\x;\theta)$ and the mixing distribution $\psi(\theta)$. Consider the variational objective with respect to the entire mixture, $\KL(\m || \p)$:
\begin{equation}\label{eqn:kl}
\begin{split}
    \KL(\m||\p) = \int_\x \m(\x) \log \frac{\m(\x)}{\p^*(\x)} \rd \x + \log \Z\, ,
\end{split}
\end{equation}
where $\Z$ is the normalizing constant of $\p^*(\x)$ and is irrelevant for constructing $\m(\x)$. Instead of (\ref{eqn:kl}), one can use the equivalent objective of maximizing the \emph{Evidence Lower BOund} or ELBO \citep{Bishop2006,Murphy2012,Blei2017}. Regardless, minimizing (\ref{eqn:kl}) or maximizing the ELBO for mixtures is intractable in general. However, as first shown by \citet{Jaakkola1998} for finite mixtures, it admits the following useful decomposition:
\begin{equation}\label{eqn:kl_mi}
\begin{split}
    \KL(\m||\p) &= \underbrace{\int_\theta \psi(\theta) \int_\x \q(\x;\theta) \log \frac{\q(\x;\theta)}{\p^*(\x)} \rd\x\,\rd\theta}_\text{(i) Expected KL} \\
    &- \underbrace{\int_\theta \psi(\theta) \int_\x \q(\x;\theta) \log \frac{\q(\x;\theta)}{\m(\x)} \rd\x\,\rd\theta}_{\text{(ii) Mutual Information } \MI[\x;\theta]}
\end{split}
\end{equation}
(dropping $\log \Z$).
The first term, (i), is the \textbf{Expected KL Divergence} for each component when the parameters are drawn from $\psi(\theta)$. This term quantifies, on average, how well the mixture components match the target distribution. In isolation, Expected KL is minimized when all components individually minimize $\KL(\q||\p)$, i.e. when $\psi(\theta) \rightarrow \delta(\theta-\thetastar)$. This tendency to concentrate $\psi(\theta)$ to the single best variational solution is balanced by the second term, (ii), which is the \textbf{Mutual Information} between $\x$ and $\theta$, which we will write $\MI[\x;\theta]$, under the joint distribution $\q(\x;\theta)\psi(\theta)$. This term should be \emph{maximized}, and, importantly, it does not depend on $\p^*(\x)$. Mutual Information is maximized when the components are as diverse as possible, which encourages the components to become narrow and to spread out over diverse regions of $\x$ \emph{regardless} of how well they agree with $\p(\x)$.
This decomposition of $\KL(\m||\p)$ into Mutual Information (between $\x$ and $\theta$) and Expected KL (between $\q$ and $\p$) is convenient because approximations to Mutual Information are well-studied, and minimizing Expected KL can leverage standard tools from VI.

\subsection{Trading Off Between Mututal Information and Expected KL}

We will refer back to this decomposition of the $\KL(\m||\p)$ objective into Expected KL (between $\q$ and $\p$) and Mutual Information (between each $\x$ and $\theta$) throughout. Figure \ref{fig:mi_kl_space} depicts a two-dimensional space with Expected KL on the x-axis and Mutual Information on the y-axis. Any given mixing distribution $\psi$ can be placed as a point in this space, but in general many $\psi$'s may map to the same point.

Sampling and VI live at extreme points in this space. Classic VI, where $\psi(\theta)=\delta(\theta-\thetastar)$, corresponds to the blue point (c), because by definition $\thetastar$ achieves the minimum possible $\KL$, and $\MI[\x;\theta]$ is zero. Classic sampling corresponds to the green point (d), with $\psi(\theta)$ placing mass only on Dirac-delta-like components, and selecting each component with probability $\p(\mu)$, where $\mu$ is the mean of $\q$ determined by $\theta$.

Towards the goal of constructing mixtures that trade-off properties of sampling and VI, we propose to view the two terms in (\ref{eqn:kl_mi}) as separate objectives that may be differently weighted, and maximizing the objective
\begin{equation}\label{eqn:weighted_optim}
    \L(\psi,\lambda) = \MI[\x;\theta] - \lambda \E_\psi\left[\KL(\q||\p)\right]
\end{equation}
for a given hyperparameter $\lambda$ with respect to the mixing distribution $\psi$. 
This objective may alternatively be viewed as the Lagrangian of a constrained optimization problem over the mixing density $\psi$, where Mutual Information is maximized subject to a constraint on Expected KL. This is a concave maximization problem with linear constraints, defining a Pareto front of solutions that each achieve a different balance between Expected KL and Mutual Information. 
In practice, maximizing Mutual Information necessitates approximations \citep{Poole2019}, so there may be good mixture approximations that are not found in practice, such as the yellow point (e) in Figure \ref{fig:mi_kl_space}. In section \ref{sec:algorithm} below, we use an approximation to Mutual Information that has the property, illustrated by the orange curve (f) in Figure \ref{fig:mi_kl_space}, of connecting VI (c) to sampling (d), controlled by varying $\lambda$. As shown on the right of Figure \ref{fig:mi_kl_space}, our method produces mixtures that behave like classic samples when $\lambda=1$, that behave like classic VI when $\lambda\rightarrow\infty$, and that exhibit intermediate behavior at intermediate values of $\lambda$.

We emphasize that this frame is quite general: any stochastic mixture can be reasoned about in terms of its Expected KL and Mutual Information, and this is a natural space in which to think about interpolating sampling and VI. A similar decomposition of $\KL(\m||\p)$ (or the ELBO) has been used by previous methods that optimize mixtures \citep{Zobay2014,Jaakkola1998,Gershman2012b,Yin2018}. The primary difference between these previous methods is how they approximate (or lower-bound) Mutual Information. In the next section, we introduce a new approximation that is particularly efficient, and is the first to our knowledge that can produce sampling-like behavior with finitely many components.

\section{Approximate objective}\label{sec:algorithm}

Maximizing Mutual Information, as is required by (\ref{eqn:weighted_optim}), is a notoriously difficult problem that arises in many domains, and there is a large collection of approximations and bounds in the literature \citep{Jaakkola1998,Brunel1998,Gershman2012b,Wei2016,Kolchinsky2017,Poole2019}. Previous work has optimized \emph{finite} mixtures by considering how each of $T$ components interacts with the other $T-1$ components, resulting in quadratic scaling with $T$ \citep{Gershman2012b,Guo2016,Miller2017,Kolchinsky2017,Yin2018,Poole2019}. Beginning instead with \emph{infinite} mixtures, we find that the local geometry of $\theta$-space is sufficient to provide an approximation to Mutual Information \emph{that can be evaluated independently for each value of $\theta$}. 

\subsection{Stam's inequality}

Mutual Information between $\x$ and $\theta$ can be written as
\begin{align}
\MI[\x;\theta] &= \H[\theta] - \E_{\m(\x)}\left[\H[\hat{\theta}|\x]\right] \nonumber \\
    &=\H[\theta] - \E_{\psi(\theta)}\big[\underbrace{\E_{\q(\x|\theta)}[\H[\hat\theta|\x]]}_{\H[\hat\theta|\theta]}\big]\label{eqn:mi_theta_hat}
\end{align}
where $\H[\theta]$ is the entropy of $\psi(\theta)$ and $\H[\hat\theta|\x]$ is the entropy of $\q(\hat\theta|\x) = \frac{\q(\x;\hat\theta)\psi(\hat\theta)}{\m(\x)}$, i.e. the distribution of \emph{inferred} $\theta$ values for a given $\x$. The second line follows simply from expanding the definition of $\m(\x)$ in the outer expectation. The term $\H[\hat\theta|\theta]$ can be thought of in terms of a statistical estimation problem: $\hat\theta$ is the ``recovered'' value of $\theta$ after passing through the ``channel'' $\x$. Bounding the error of such estimators is a well-studied problem in statistics.

From (\ref{eqn:mi_theta_hat}), a lower-bound on Mutual Information can be derived from an \emph{upper bound} on $\H[\hat\theta|\theta]$ for each $\theta$. 
For this, we draw inspiration from Stam's inequality \citep{Stam1959,Dembo1991,Wei2016}, which states
\begin{equation}\label{eqn:stams}
    \H[\hat\theta|\theta] \leq \frac{1}{2}\log\left|2\pi e \FIM(\theta)^{-1}\right| \, ,
\end{equation}
where $| \cdot |$ is a determinant, and $\FIM(\theta)$ is the Fisher Information Matrix, defined as
\begin{align*}
    \FIM(\theta)_{ij} = -\E_{\q(\x;\theta)}\left[\frac{\partial^2}{\partial \theta_i\partial\theta_j}\log\q(\x;\theta)\right]\, .
\end{align*}
The Fisher Information Matrix is also the local metric on the \emph{statistical manifold} with coordinates $\theta$ \citep{Amari2016}; it is used to quantify how ``distinguishable'' $\theta$ is from $\theta+d\theta$. Note that (\ref{eqn:stams}) can be viewed as the entropy of a Gaussian approximation to $\q(\hat\theta|\x)$ with precision matrix $\FIM(\theta)$; this approximation is most accurate when $\q(\x;\theta)$ itself is narrow and approximately Gaussian \citep{Wei2016}. 

Combining (\ref{eqn:mi_theta_hat}) and (\ref{eqn:stams}), we propose to use
\begin{equation}\label{eqn:mi_stams}
    \MI_\FIM[\x;\theta] \equiv \H[\theta] - \frac{1}{2}\E_{\psi(\theta)}\left[\log\left|2\pi e \FIM(\theta)^{-1}\right|\right]
\end{equation}
as a proxy for the intractable $\MI[\x;\theta]$ in (\ref{eqn:weighted_optim}). 

Note that $\MI_\FIM[\x;\theta]$ is not strictly a \emph{bound} on $\MI[\x;\theta]$, but may be seen as an \emph{approximation} to it \citep{Wei2016}. Briefly, this is because the original Stam's inequality, as stated in (\ref{eqn:stams}), assumes $\theta$ is a scalar location parameter, and assumes the high-precision limit where $\q(\hat\theta|\x)$ is well-approximated by a Gaussian. Despite this, $\MI_\FIM[\x;\theta]$ is well-suited for our purposes, since (i) it leads to a remarkably simple and easy to implement expression for $\psi(\theta)$ below; (ii) we can prove that it leads to sampling when $\lambda=1$ and VI when $\lambda\rightarrow\infty$; and (iii) the inequality in (\ref{eqn:stams}) is nonetheless likely to be strict, since we neglect the prior information contained in $\psi(\theta)$ when estimating $\hat\theta$ and therefore over-estimate the entropy.\footnote{By analogy to the Bayesian Cram{\' e}r-Rao bound \citep{Gill1995,Fauss2021}, a tighter variant of (\ref{eqn:stams}) could be derived that takes into account the prior, though possibly at the expense of added complexity; we leave this to future work.}

\subsection{Closed-form mixing distribution}

Substituting $\MI_\FIM[\x;\theta]$ for $\MI[\x;\theta]$ in (\ref{eqn:weighted_optim}) gives the following approximate objective,
\begin{equation}\label{eqn:weighted_optim_f}
    \L_\FIM(\psi,\lambda) = \H[\theta] + \E_\psi\left[\frac{1}{2}\log|\FIM| - \lambda\, \KL(\q||\p^*)\right]
\end{equation}
having dropped additive constants and using $\log|\FIM^{-1}|=-\log|\FIM|$.
This now resembles a maximum-entropy problem with an expected-value constraint, which has the following simple closed-form solution:
\begin{equation}\label{eqn:log_psi_stams}
    \log\psi(\theta) = \frac{1}{2}\log|\FIM(\theta)| - \lambda\, \KL(\q(\x;\theta)||\p^*(\x))
\end{equation}
again dropping additive constants. Equation (\ref{eqn:log_psi_stams}) is strikingly simple, and amenable to many existing MCMC sampling methods for drawing samples of $\theta$ from $\psi$. 

Despite being derived from an approximation to our original objective, (\ref{eqn:log_psi_stams}) nonetheless contains both sampling and VI as special cases. As $\lambda\rightarrow\infty$, the $\KL$ term dominates and $\psi(\theta)$ concentrates to $\delta(\theta-\thetastar)$, reproducing VI. When $\lambda=1$, this mixing distribution also corresponds to ``sampling'' in the following sense:
\begin{definition}[Sampling]\label{def:sampling}
A stochastic mixture, defined by the component family $\q(\x;\theta)$ and mixing distribution $\psi(\theta)$, is considered to be ``sampling'' if (i) it is \textbf{unbiased} in the limit of infinitely many components, i.e. $\m(\x) \rightarrow \p(\x)$; and, (ii) it consists of \textbf{non-overlapping components}. That is, for small values of of $0 < \epsilon \ll 1$, wherever $\q(\x;\theta_i) > \epsilon$, with high probability, $\q(\x;\theta_j) < \epsilon$, for all pairs $\theta_i, \theta_j$ drawn independently from $\psi(\theta)$.
\end{definition}
Lemma \ref{lem:sampling_approximate_solution} in Appendix \ref{app:sampling} establishes that $\psi(\theta)$ with $\lambda=1$ leads to sampling as defined here, assuming mixture components $\q$ are Gaussian. However, we conjecture that sampling arises from a broader class of $\q$ components as well.

\subsection{Implementation}

We implemented (\ref{eqn:log_psi_stams}) in Stan \citep{Carpenter2017}, an open-source framework for probabilistic models and approximate inference algorithms. We sampled $\theta$ from $\psi(\theta)$ using Stan's default implementation of the No U-Turn Sampler (NUTS)  \citep{Hoffman2014}, but we emphasize that samples can be drawn from (\ref{eqn:log_psi_stams}) using any existing sampling method. All comparisons to existing methods were with Stan's built-in NUTS sampler (over $\x$) and its built-in mean-field VI \citep{Kucukelbir2017}.

\section{Navigating Bias/Variance Trade-Offs For Finite $T$}\label{sec:bias_variance}

\begin{figure}[t]
    \centering
    \includegraphics[width=\linewidth]{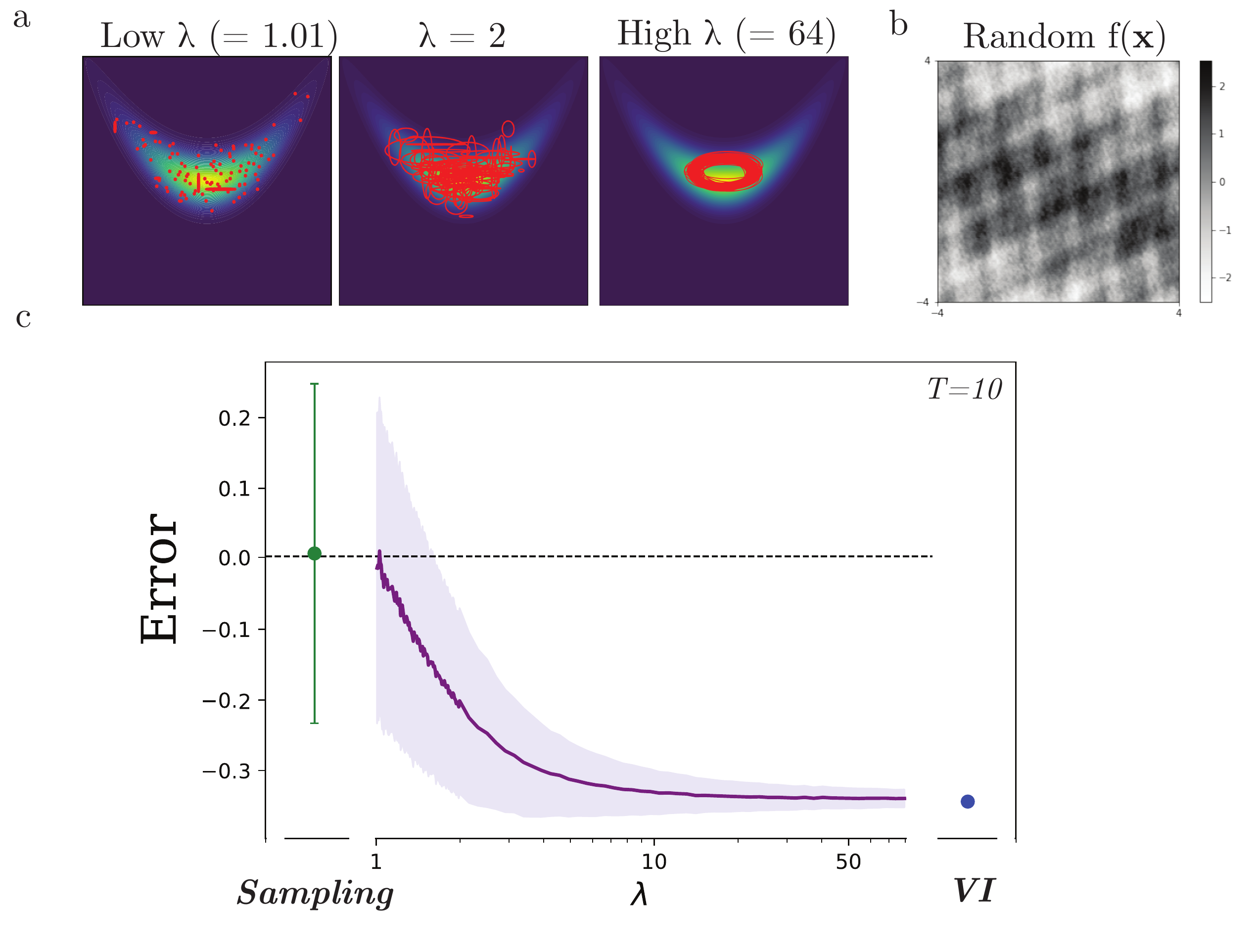}
    \caption{$\lambda$ controls a bias/variance tradeoff, interpolating between sampling and VI. 
    \textbf{a)} For an example 2D distribution (the banana distribution), we set $\q$ to Gaussian with diagonal covariance and sampled $\theta\sim\psi(\theta)$ using NUTS (see Appendix B.2 for sampling details). 
    \textbf{b)} We selected $f(\x)$ as a random mixture of sinusoids at different frequencies. We then calculated the bias and variance of computing $\E_{\m_T}[f(\x)]$.
    \textbf{c)} The green point and error bars (``Sampling'') indicate the estimated value of $\E_{\p(\x)}[f(\x)]$ and its variance using NUTS to draw samples of $\x$. The blue point (``VI'') shows the value of $\E_{\q(\x;\thetastar)}[f(\x)]$ using Stan's built-in VI. Our method is shown in the middle across a range of $\lambda$ values. Low $\lambda$ provides unbiased but high variance estimators, while high $\lambda$ provides a bias near that of standard VI and a vanishing variance. In panel (c), we used $T=10$ independent samples for both classic NUTS and our method.
    }
    \label{fig:bias_variance}
\end{figure}

\begin{figure*}[ht]
    \centering
    \includegraphics[width=\textwidth]{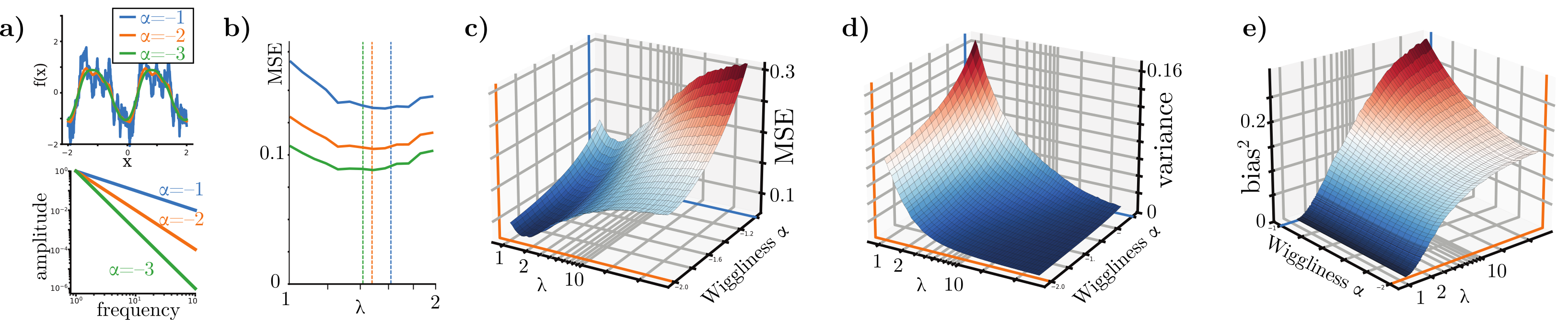}
    \caption{Flexibly trading bias for variance matters for integrating functions. 
    \textbf{a)} We generated random functions of a specified smoothness by varying the decay of its power spectrum while randomizing phase. 
    \textbf{b)} The $\lambda$ with the smallest MSE for a fixed number samples $T=100$ depends on the integrand’s smoothness.
    \textbf{c)} Surface plots of MSE for $T=100$ samples, varying $\lambda$ and $\alpha$. 
    \textbf{d)} Variance is higher for smaller $\lambda$ and more wiggly integrands. 
    \textbf{e)} Bias vanishes near $\lambda=1$.
    }
    \label{fig:wiggliness}
\end{figure*}

\subsection{Reducing Mean Squared Error (MSE)}

In this section, we expound the sense in which our method ``interpolates'' sampling and VI in terms of bias and variance, both analytically and empirically. In our experiments, we quantify bias and variance in terms of the Mean Squared Error (MSE) of the expectation of an arbitrary $f(\x)$ using a random mixture of $T$ components, $\m_T(\x)$. In Figure \ref{fig:bias_variance}, we show empirically that by increasing $\lambda$ one can interpolate between the zero bias but high variance solution, equivalent to sampling, and the zero variance but high bias solution, equivalent to VI. Between these extremes, our method smoothly interpolates both bias and variance.

To show this empirically requires choosing a class of functions $f(\x)$. We construct a random smooth function by discrete Fourier synthesis. Specifically, we select a series of sinusoid plane waves in the space of $\x$ with increasing frequency $\omega$, random directions $\randomt$ and phase $\phi$, such that $f(\x) = \sum_{\omega=1}^N a_\omega \sin(\omega \randomt^\top\x + \phi_\omega)$.
The amplitudes $a_\omega$ are set according to a power law: $a_\omega=\omega^{-\alpha}$. An example of $f(\x)$ is shown in Fig. \ref{fig:bias_variance}b for $\alpha=-1$, and $\alpha$ is varied in Fig \ref{fig:wiggliness}. Adjusting $\alpha$ allows flexibly setting the ``wiggliness'' of the synthesized function \citep{stein2011fourier}.

We also tested our algorithm on three reference problems from posteriordb \citep{posteriordb}, now evaluating a large set of random $f$s, defined on the space of each model's unconstrained parameters, with $\alpha=-1$ (Figure \ref{fig:posteriordb}). The conclusion is similar: across many random $f$s, our algorithm performs on average as well as or better than both sampling (by reducing variance) and VI (by reducing bias).

\subsection{Considerations for selecting $\lambda$}

A first practical consideration for the choice of $\lambda$ is the particular function $f(\x)$ to be integrated. Since MSE can be decomposed into the sum of squared bias and variance, the value of $\lambda$ that minimizes MSE occurs when $\frac{\partial\textrm{Bias}^2}{\partial\lambda}=-\frac{\partial\textrm{Var}}{\partial\lambda}$. Any factor that increases the variance but not the bias of an estimate for a fixed number of components $T$ will push the optimal $\lambda$ towards higher values. 

One such factor is the smoothness of $f(\x)$. Classic sampling can have problematically high variance when $f(\x)$ is very jagged, as single points are not very representative of the surrounding function. Intuitively, then, we should expect that higher $\lambda$ (more VI-like mixtures) is preferred when $f(\x)$ is more ``wiggly.'' To show this, we generated a random function with varying smoothness, integrated it over random mixtures approximating the 2D banana distribution, and plotted the resulting MSE, bias, and variance (Fig. \ref{fig:wiggliness}). We adjusted smoothness by varying the power law decay, $\alpha$, for a fixed set of phases and wave directions. At any value of $\lambda$, variance can be seen to increase as $f$ is made more wiggly. With all else held equal, it is better to trade some variance for bias when the integrand changes quickly with $\x$. 

Another factor that affects the optimal $\lambda$ is the computational budget. If time allows a large number of $T$ to be sampled, the optimal $\lambda$ will approach $1$ with a speed that depends on the particular problem (specifically, on $\frac{\partial\textrm{Bias}^2}{\partial\lambda}$). In our experiments we set a fixed $T$ to demonstrate our algorithm's properties. However, if the number of components is not known in advance, a practitioner may also decrease $\lambda$ adaptively over time as sampling continues.

\subsection{Analytical results}

While the MSE of the expected value of some $f(\x)$ is a useful way to compare approximate inference methods, it depends on the somewhat arbitrary choice of $f$, and in practice, the $f$'s of interest are often not known at the time of inference. This motivates using the following alternative definition of error that is independent of $f$ and closely related to the variational objective of minimizing $\KL$ divergence:
\begin{equation}\label{eqn:kl_error}
\begin{split}
    \text{KL error} = \E[\KL(\m_T(\x)||\p(\x))] = \\
    \underbrace{\KL(\m(\x)||\p(\x))}_\text{KL bias} + \underbrace{\E\left[\KL(\m_T(\x)||\m(\x))\right]}_\text{KL variance} \, .
\end{split}
\end{equation}
That is, \textbf{KL bias} is the $\KL$ divergence from the infinite mixture $\m(\x)$ to the true distribution, and \textbf{KL variance} is the average $\KL$, over realizations of $T$ independent mixture components, from $\m_T(\x)$ to the infinite mixture $\m(\x)$. Note that KL bias is identical to the infinite-mixture objective we started with in (\ref{eqn:kl_mi}).

The following theorem establishes that for all finite $T$, we can always reduce the KL error, relative to sampling, using some $\lambda > 1$.
\begin{theorem}[Improve on sampling]\label{thm:improve_sampling}
If a mixture is sampling as in Definition \ref{def:sampling}, then $\frac{\rd}{\rd\lambda}\text{KL bias}=0$ and $\frac{\rd}{\rd\lambda}\text{KL variance} < 0$. Thus, $\frac{\rd}{\rd\lambda}\text{KL error}<0$.
\end{theorem}
This theorem establishes the intuitive result that the variance of sampling can be reduced, minimally impacting its bias, by replacing samples with narrow mixture components. 
Importantly, Theorem \ref{thm:improve_sampling} is based on how $\psi(\theta)$ changes with $\lambda$ when using the closed-form expression for $\psi$ we derived based on the approximate $\L_\FIM$ objective. Because the theorem is phrased in conditional terms (``if the mixture is sampling, then...''), we must further show that both conditions of ``sampling'' (Definition \ref{def:sampling}) are met when $\lambda=1$. This is proved in Lemma \ref{lem:sampling_approximate_solution} in Appendix \ref{app:sampling} for Gaussian components, though we suspect it holds for other component families as well.

We can also improve on VI using stochastic mixtures. However, this result is slightly more subtle, as there are three cases where one should expect VI to be optimal. First, if $\q$ is in the same family as $\p$, then $\q(\x;\thetastar)=\p(\x)$, then is no benefit to increasing $T$, and reducing $\lambda$ only adds variance. Second, if $T$ is small -- in the most extreme case, if $T=1$ -- then reducing $\lambda$ will again only add variance without reducing bias. Third, if $\p$ is lighter-tailed than $\q(\x;\thetastar)$, then a mixture of nearby $\q$s will add variance to $\m$ \citep{Lindsay1983a}, making the match to $\p$ worse. With these three exceptions in mind, the following theorem establishes conditions where we expect to reduce KL error relative to VI by using a large but finite $\lambda < \infty$.
\begin{theorem}[Improve on VI]\label{thm:improve_vi}
Assume that $\p(\x)$ is heaver-tailed than $\q(\x;\thetastar)$ and that $\lambda$ is large. Then, there exists some finite $T_0>1$ such that for all $T\geq T_0$, $\frac{\rd}{\rd\lambda}\text{KL error}>0$.
Proof: see Appendix \ref{app:vi}.
\end{theorem}
Note that this result depends on an additional conjecture that relates the curvature in parameter space of $\KL(\q||\p)$ to the curvature of $\KL(\q||\q^*)$ that we believe holds as long as $\p$ is heavier-tailed than $\q$. For details, see Appendix \ref{app:vi}.

\section{DISCUSSION}\label{sec:discussion}

\paragraph{Summary:} Our work provides a new perspective on the relationship between the two dominant frameworks for approximate inference -- sampling and VI -- by viewing both as special cases of inference using a broader class of stochastic mixtures. Our main theoretical contribution is the framework shown in Figure \ref{fig:mi_kl_space}, where mixtures that ``interpolate'' sampling and VI are analyzed in terms of how they trade off Mutual Information and Expected KL. We then derived an easy-to-use method based on an approximation to Mutual Information that uses the local geometry of the space of variational parameters. To demonstrate the ease and effectiveness of our method, we implemented it in the popular Stan language and demonstrated using a small set of reference problems how we ``interpolate'' sampling and VI by varying a single parameter, $\lambda$. Finally, we showed why such an intermediate inference scheme is useful in practice. On one hand, we proved that it is always possible to improve on classic sampling ($\lambda=1$) by increasing $\lambda$: our method provably reduces the variance of sampling while minimally impacting its bias. On the other hand,  our method provably reduces the bias of VI under certain conditions (and improves overall error if the number of mixture components is sufficiently large).

\paragraph{Time and space complexity:}
By approximating Mutual Information using only \emph{local} geometric information in (\ref{eqn:mi_stams}), in our method each component can be selected independently of the others. This means we can select and evaluate $T$ components in $\mathcal{O}(T)$ time and either $\mathcal{O}(T)$ space (if all are stored) or $\mathcal{O}(1)$ space (if components are evaluated online) -- identical to traditional MCMC sampling algorithms. Further, we can run independent chains sampling $\theta \sim \psi(\theta)$ for a constant factor speedup. This improves on past work using mixture approximations, which incurred $\mathcal{O}(T^2)$ time and $\mathcal{O}(T)$ space complexity, since the optimization problem for the $T$th component depends on the location of the other $T-1$ components, all of which must be in memory at once \citep{Jaakkola1998,Gershman2012b,Salimans2015,Guo2016,Miller2017,Acerbi2018,Yin2018} (but the $\mathcal{O}(T^2)$ complexity may be hardware-accelerated). 

\paragraph{Related Work:} The trade-offs between sampling and VI are well-studied, and many methods have been proposed to ``close the gap'' between them (see \citep{Angelino2016,Zhang2019} for general reviews). Like these other methods, we aim to provide good approximations with high computational efficiency and low variance.

There are many methods that use mixture models to reduce the bias of variational inference.
Theorem \ref{thm:improve_vi} shows that our method only ``beats'' classic VI when $T>T_0$ for some finite but potentially large $T_0$. This is the price we pay for drawing mixture components stochastically \citep{Yin2018}. When a mixture of $T$ components is \emph{optimized} rather than \emph{sampled}, bias is reduced and variance remains near zero, as in previous work \citep{Jaakkola1998,Gershman2012b,Zobay2014,Guo2016,Miller2017}, but in previous work this optimization has incurred a $\mathcal{O}(T^2)$ cost while our method is $\mathcal{O}(T)$ and can be further parallelized. Further, with some notable exceptions \citep{Anaya-Izquierdo2007,Salimans2015}, most mixture VI methods make strong assumptions about the family of components \citep{Jaakkola1998,Gershman2012b,Acerbi2018,Miller2017}. Our framework and method is somewhat agnostic to the family of $\q$, though we have only rigorously proved that is asymptotically unbiased when using Gaussian components.

Many methods use sampling in the service of variational inference, or vice versa, but do not provide a unifying approach to both. These typically use the samples to compute expectations used to update a variational approximation \citep{Acerbi2018,Miller2017,Kucukelbir2017}, rather than to generate the mixture components themselves. 

There is also a large number of sampling approaches that aim to improve the efficiency of sampling by reducing its variance at the cost of some bias. Some of these use variational approaches as proposal distributions, but ultimately the posterior is approximated by a set of (possibly weighted) samples of the latent variables \citep{deFreitas2001,Korattikara2013,Ma2015,Zhang2021}. By expanding each sample to a distribution, our approach allows each sample to cover more space with less variance and greater efficiency \citep{Nalisnick2017}.

Despite some high-level similarities to other approaches, our framework is unusual in approximating the posterior by a sampled mixture of variational approximations. The Mixture Kalman filter \citep{chen2000mixture} is a special case of this, which uses a sampled mixture of Gaussians, each constructed as a Kalman filter. A related approach is to \emph{optimize} a parameterized function that generates mixture components \citep{Salimans2015, wolf2016variational,Yin2018}, and generative diffusion models can also be seen as a case of this approach \citep{sohl2015deep, ho2020denoising}. Our work differs in that we derived a closed-form mixing distribution that requires no additional learning or optimization and that is readily implemented in existing inference software (Stan, \citep{Carpenter2017}).

\paragraph{Limitations and future work:} Using $\MI_\FIM[\x;\theta]$ to approximate $\MI[\x;\theta]$ reduces the generality of our method, since the former is most appropriate for narrow and Gaussian-like components \citep{Wei2016}. Incorporating prior information from $\psi(\theta)$ into this bound, generalizing to other kinds of components, or even starting with alternative bounds on $\MI[\x;\theta]$ are all interesting avenues for future work. Another limitation of our theory is that our proof of Theorem \ref{thm:improve_vi} depends on a conjecture.

We currently only study mixtures with $T$ \emph{independent} mixture components without taking into account the cost of producing independent samples of $\theta$. In reality, this cost depends on the quality of the sampler, warm-up and burn-in time, and a potentially large number of calls to $\log\p(\x)$ \citep{Zhang2021}. Further, $\lambda$ dramatically changes the shape of $\log\psi(\theta)$, which may affect the efficiency of the sampler -- we mitigated this slightly by scaling the mass parameter of NUTS with $\lambda$.

We have so far considered $\lambda$ to be constant for a run of our algorithm, and this can lead to asymptotic bias even when $T$ is large. A simple adjustment to make our method effective at both small and large $T$ would be to decay $\lambda$ as $T$ grows, but note that this may require adapting the sampler parameters on the fly. Our method also requires evaluating $\KL(\q||\p)$ many times per sample of $\theta$. This could be made more efficient by adapting the number of Monte Carlo evaluations (fewer samples from $\q$ are sufficient when $\lambda$ is low and components are narrow), by accounting for stochastic likelihood evaluations \citep{Ma2015}, or by extending our method to mean-field message-passing \citep{Jaakkola1998}, where $\nabla_\theta\KL(\q||\p)$ can be computed in closed form \citep{Hoffman2013}.

\begin{acknowledgements}
We thank Emmett Wyman and Roozbeh Farhoudi for helpful discussions early on, and Konrad Kording for suggestions on writing and presentation. Daniel Lee's advice was indispensable for getting our algorithm to run in Stan.
\end{acknowledgements}

\bibliography{references-mendeley-group,additional-references}

\begin{thebibliography}{48}
\providecommand{\natexlab}[1]{#1}
\providecommand{\url}[1]{\texttt{#1}}
\expandafter\ifx\csname urlstyle\endcsname\relax
  \providecommand{\doi}[1]{doi: #1}\else
  \providecommand{\doi}{doi: \begingroup \urlstyle{rm}\Url}\fi

\bibitem[Acerbi(2018)]{Acerbi2018}
Luigi Acerbi.
\newblock {Variational Bayesian Monte Carlo}.
\newblock \emph{Advances in Neural Information Processing Systems}, 2018.

\bibitem[Amari(2016)]{Amari2016}
S~Amari.
\newblock \emph{{Information Geometry and Its Applications}}.
\newblock Applied Mathematical Sciences. Springer Japan, 2016.
\newblock ISBN 9784431559788.
\newblock URL \url{https://books.google.com/books?id=UkSFCwAAQBAJ}.

\bibitem[Anaya-Izquierdo and Marriott(2007)]{Anaya-Izquierdo2007}
Karim Anaya-Izquierdo and Paul Marriott.
\newblock {Local mixture models of exponential families}.
\newblock \emph{Bernoulli}, 13\penalty0 (3):\penalty0 623--640, 2007.
\newblock ISSN 1350-7265.
\newblock \doi{10.3150/07-BEJ6170}.
\newblock URL \url{http://projecteuclid.org/euclid.bj/1186503479}.

\bibitem[Angelino et~al.(2016)Angelino, Johnson, and Adams]{Angelino2016}
Elaine Angelino, Matthew~James Johnson, and Ryan~P. Adams.
\newblock {Patterns of Scalable Bayesian Inference}.
\newblock \emph{Foundations and Trends in Machine Learning}, 9\penalty0
  (2-3):\penalty0 119--247, 2016.
\newblock \doi{10.1561/2200000052}.

\bibitem[Besag et~al.(1995)Besag, Green, Higdon, and Mengersen]{Besag1995}
Julian Besag, Peter Green, David Higdon, and Kerrie Mengersen.
\newblock {Bayesian computation and stochastic systems}.
\newblock \emph{Statistical Science}, 10\penalty0 (1):\penalty0 3--44, 1995.
\newblock ISSN 08834237.
\newblock \doi{10.1214/ss/1177010123}.

\bibitem[Bishop(2006)]{Bishop2006}
Christopher~M Bishop.
\newblock {Pattern Recognition and Machine Learning}.
\newblock \emph{Pattern Recognition}, page 738, 2006.
\newblock ISSN 10179909.
\newblock \doi{10.1117/1.2819119}.
\newblock URL \url{http://www.library.wisc.edu/selectedtocs/bg0137.pdf}.

\bibitem[Blei et~al.(2017)Blei, Kucukelbir, and Mcauliffe]{Blei2017}
David~M. Blei, Alp Kucukelbir, and Jon~D Mcauliffe.
\newblock {Variational Inference: A Review for Statisticians}.
\newblock \emph{arXiv}, pages 1--41, 2017.

\bibitem[Braverman and Bhowmick(2011)]{Braverman2011}
Mark Braverman and Abhishek Bhowmick.
\newblock Convexity/concavity of mutual information, September 2011.
\newblock URL
  \url{https://www.cs.princeton.edu/courses/archive/fall11/cos597D/L04.pdf}.

\bibitem[Brunel and Nadal(1998)]{Brunel1998}
Nicolas Brunel and Jean~Pierre Nadal.
\newblock {Mutual Information, Fisher Information, and Population Coding}.
\newblock \emph{Neural Computation}, 10\penalty0 (7):\penalty0 1731--1757,
  1998.
\newblock ISSN 08997667.
\newblock \doi{10.1162/089976698300017115}.

\bibitem[Carpenter et~al.(2017)Carpenter, Gelman, Hoffman, Lee, Goodrich,
  Betancourt, Brubaker, Guo, Li, and Riddell]{Carpenter2017}
Bob Carpenter, Andrew Gelman, Matthew~D. Hoffman, Daniel Lee, Ben Goodrich,
  Michael Betancourt, Marcus~A. Brubaker, Jiqiang Guo, Peter Li, and Allen
  Riddell.
\newblock {Stan: A probabilistic programming language}.
\newblock \emph{Journal of Statistical Software}, 76\penalty0 (1), 2017.
\newblock ISSN 15487660.
\newblock \doi{10.18637/jss.v076.i01}.

\bibitem[Chen and Liu(2000)]{chen2000mixture}
Rong Chen and Jun~S Liu.
\newblock Mixture kalman filters.
\newblock \emph{Journal of the Royal Statistical Society: Series B (Statistical
  Methodology)}, 62\penalty0 (3):\penalty0 493--508, 2000.

\bibitem[de~Freitas et~al.(2001)de~Freitas, H{\o}jen-S{\o}rensen, Jordan, and
  Russel]{deFreitas2001}
Nando de~Freitas, Pedro H{\o}jen-S{\o}rensen, Michael~I. Jordan, and Stuart
  Russel.
\newblock {Variational MCMC}.
\newblock \emph{Uncertainty in Artificial Intelligence}, 2001.

\bibitem[Dembo et~al.(1991)Dembo, Cover, and Thomas]{Dembo1991}
Amir Dembo, Thomas~M. Cover, and Joy~A. Thomas.
\newblock {Information Theoretic Inequalities}.
\newblock \emph{IEEE Transactions on Information Theory}, 37\penalty0
  (6):\penalty0 1501--1518, 1991.
\newblock ISSN 15579654.
\newblock \doi{10.1109/18.104312}.

\bibitem[Fau{\ss} et~al.(2021)Fau{\ss}, Dytso, and Poor]{Fauss2021}
Michael Fau{\ss}, Alex Dytso, and H.~Vincent Poor.
\newblock {A variational interpretation of the Cram{\'{e}}r–Rao bound}.
\newblock \emph{Signal Processing}, 182, 2021.
\newblock ISSN 01651684.
\newblock \doi{10.1016/j.sigpro.2020.107917}.

\bibitem[Gershman et~al.(2012)Gershman, Hoffman, and Blei]{Gershman2012b}
Samuel~J. Gershman, Matthew~D. Hoffman, and David~M. Blei.
\newblock {Nonparametric Variational Inference}.
\newblock \emph{Proceedings of the 29th International Conference on Machine
  Learning}, pages 235--242, 2012.
\newblock ISSN 0899-7667.
\newblock \doi{10.1162/089976699300016331}.
\newblock URL \url{https://icml.cc/Conferences/2012/papers/360.pdf}.

\bibitem[Gill and Levit(1995)]{Gill1995}
Richard~D. Gill and Boris~Y. Levit.
\newblock {Applications of the van Trees inequality: A Bayesian
  Cram{\'{e}}r-Rao bound}.
\newblock \emph{Bernoulli}, 1\penalty0 (1):\penalty0 59--79, 1995.
\newblock URL \url{https://www.jstor.org/stable/3318681}.

\bibitem[Guo et~al.(2016)Guo, Wang, Fan, Broderick, and Dunson]{Guo2016}
Fangjian Guo, Xiangyu Wang, Kai Fan, Tamara Broderick, and David~B. Dunson.
\newblock {Boosting Variational Inference}.
\newblock \emph{arXiv}, 2016.
\newblock URL \url{http://arxiv.org/abs/1611.05559}.

\bibitem[Harris et~al.(2020)Harris, Millman, van~der Walt, Gommers, Virtanen,
  Cournapeau, Wieser, Taylor, Berg, Smith, Kern, Picus, Hoyer, van Kerkwijk,
  Brett, Haldane, del R{\'{i}}o, Wiebe, Peterson, G{\'{e}}rard-Marchant,
  Sheppard, Reddy, Weckesser, Abbasi, Gohlke, and Oliphant]{numpy}
Charles~R. Harris, K.~Jarrod Millman, St{\'{e}}fan~J. van~der Walt, Ralf
  Gommers, Pauli Virtanen, David Cournapeau, Eric Wieser, Julian Taylor,
  Sebastian Berg, Nathaniel~J. Smith, Robert Kern, Matti Picus, Stephan Hoyer,
  Marten~H. van Kerkwijk, Matthew Brett, Allan Haldane, Jaime~Fern{\'{a}}ndez
  del R{\'{i}}o, Mark Wiebe, Pearu Peterson, Pierre G{\'{e}}rard-Marchant,
  Kevin Sheppard, Tyler Reddy, Warren Weckesser, Hameer Abbasi, Christoph
  Gohlke, and Travis~E. Oliphant.
\newblock Array programming with {NumPy}.
\newblock \emph{Nature}, 585\penalty0 (7825):\penalty0 357--362, September
  2020.
\newblock \doi{10.1038/s41586-020-2649-2}.
\newblock URL \url{https://doi.org/10.1038/s41586-020-2649-2}.

\bibitem[Ho et~al.(2020)Ho, Jain, and Abbeel]{ho2020denoising}
Jonathan Ho, Ajay Jain, and Pieter Abbeel.
\newblock Denoising diffusion probabilistic models.
\newblock \emph{arXiv preprint arXiv:2006.11239}, 2020.

\bibitem[Hobert and Casella(1996)]{Hobert1996}
James~P. Hobert and George Casella.
\newblock {The Effect of Improper Priors on Gibbs Sampling in Hierarchical
  Linear Mixed Models}.
\newblock \emph{Journal of the American Statistical Association}, 91\penalty0
  (436):\penalty0 1461--1473, 1996.
\newblock ISSN 1537274X.
\newblock \doi{10.1080/01621459.1996.10476714}.

\bibitem[Hoffman and Gelman(2014)]{Hoffman2014}
Matthew~D Hoffman and Andrew Gelman.
\newblock {The No-U-Turn Sampler: Adaptively Setting Path Lengths in
  Hamiltonian Monte Carlo}.
\newblock \emph{Journal of Machine Learning Research}, 15:\penalty0 1351--1381,
  2014.

\bibitem[Hoffman et~al.(2013)Hoffman, Blei, Wang, and Paisley]{Hoffman2013}
Matthew~D. Hoffman, David~M. Blei, Chong Wang, and John Paisley.
\newblock {Stochastic variational inference}.
\newblock \emph{Journal of Machine Learning Research}, 14:\penalty0 1303--1347,
  2013.
\newblock ISSN 1532-4435.
\newblock \doi{citeulike-article-id:10852147}.
\newblock URL \url{http://arxiv.org/abs/1206.7051}.

\bibitem[Hunter(2007)]{matplotlib}
John~D. Hunter.
\newblock Matplotlib: A 2d graphics environment.
\newblock \emph{Computing in Science Engineering}, 9\penalty0 (3):\penalty0
  90--95, 2007.
\newblock \doi{10.1109/MCSE.2007.55}.

\bibitem[Jaakkola and Jordan(1998)]{Jaakkola1998}
Tommi~S. Jaakkola and Michael~I. Jordan.
\newblock {Improving the Mean Field Approximation via the Use of Mixture
  Distributions}.
\newblock In Michael~I. Jordan, editor, \emph{Learning in Graphical Models}.
  Kluwer Academic Publishers, 1998.

\bibitem[Kolchinsky and Tracey(2017)]{Kolchinsky2017}
Artemy Kolchinsky and Brendan~D. Tracey.
\newblock {Estimating mixture entropy with pairwise distances}.
\newblock \emph{Entropy}, 19\penalty0 (7):\penalty0 1--17, 2017.
\newblock ISSN 10994300.
\newblock \doi{10.3390/e19070361}.

\bibitem[Korattikara et~al.(2014)Korattikara, Chen, and
  Welling]{Korattikara2013}
Anoop Korattikara, Yutian Chen, and Max Welling.
\newblock {Austerity in MCMC Land: Cutting the Metropolis-Hastings Budget}.
\newblock \emph{International Conference on Machine Learning}, 32\penalty0
  (1):\penalty0 181--189, 2014.
\newblock URL \url{http://arxiv.org/abs/1304.5299}.

\bibitem[Kucukelbir et~al.(2017)Kucukelbir, Blei, Gelman, Ranganath, and
  Tran]{Kucukelbir2017}
Alp Kucukelbir, David~M. Blei, Andrew Gelman, Rajesh Ranganath, and Dustin
  Tran.
\newblock {Automatic Differentiation Variational Inference}.
\newblock \emph{Journal of Machine Learning Research}, 18:\penalty0 1--45,
  2017.
\newblock ISSN 15337928.

\bibitem[Lindsay(1983)]{Lindsay1983a}
Bruce~G. Lindsay.
\newblock {The Geometry of Mixture Likelihoods: A General Theory}.
\newblock \emph{The Annals of Statistics}, 11\penalty0 (1):\penalty0 86--94,
  1983.

\bibitem[Ma et~al.(2015)Ma, Chen, and Fox]{Ma2015}
Yi-An Ma, Tianqi Chen, and Emily~B. Fox.
\newblock {A Complete Recipe for Stochastic Gradient MCMC}.
\newblock \emph{Advances in Neural Information Processing Systems}, pages
  1--16, 2015.
\newblock ISSN 10495258.
\newblock URL \url{http://arxiv.org/abs/1506.04696}.

\bibitem[Magnusson et~al.(2021)Magnusson, B{\"{u}}rkner, and
  Vehtari]{posteriordb}
M.~Magnusson, Paul-Christian B{\"{u}}rkner, and Aki Vehtari.
\newblock {posteriordb: A database of Bayesian posterior inference}, 2021.
\newblock URL \url{https://github.com/stan-dev/posteriordb}.

\bibitem[Miller et~al.(2017)Miller, Foti, and Adams]{Miller2017}
Andrew~C. Miller, Nicholas~J. Foti, and Ryan~P. Adams.
\newblock {Variational Boosting: Iteratively Refining Posterior
  Approximations}.
\newblock \emph{arXiv}, 2017.
\newblock URL \url{http://arxiv.org/abs/1611.06585}.

\bibitem[Murphy(2012)]{Murphy2012}
Kevin~P. Murphy.
\newblock \emph{{Machine Learning: A Probabilistic Perspective}}.
\newblock The MIT Press, Cambridge, MA, 2012.

\bibitem[Nalisnick and Smyth(2017)]{Nalisnick2017}
Eric Nalisnick and Padhraic Smyth.
\newblock {Variational Inference with Stein Mixtures}.
\newblock \emph{NIPS2017 (Workshop)}, 2017.
\newblock ISSN 00368075.
\newblock \doi{10.1126/science.1070850}.
\newblock URL
  \url{https://www.ics.uci.edu/$\sim$enalisni/AABI_paper30-Stein_Mixtures.pdf}.

\bibitem[Paszke et~al.(2019)Paszke, Gross, Massa, Lerer, Bradbury, Chanan,
  Killeen, Lin, Gimelshein, Antiga, Desmaison, Kopf, Yang, DeVito, Raison,
  Tejani, Chilamkurthy, Steiner, Fang, Bai, and Chintala]{pytorch}
Adam Paszke, Sam Gross, Francisco Massa, Adam Lerer, James Bradbury, Gregory
  Chanan, Trevor Killeen, Zeming Lin, Natalia Gimelshein, Luca Antiga, Alban
  Desmaison, Andreas Kopf, Edward Yang, Zachary DeVito, Martin Raison, Alykhan
  Tejani, Sasank Chilamkurthy, Benoit Steiner, Lu~Fang, Junjie Bai, and Soumith
  Chintala.
\newblock Pytorch: An imperative style, high-performance deep learning library.
\newblock In \emph{Advances in Neural Information Processing Systems 32}, pages
  8024--8035. Curran Associates, Inc., 2019.
\newblock URL
  \url{http://papers.neurips.cc/paper/9015-pytorch-an-imperative-style-high-performance-deep-learning-library.pdf}.

\bibitem[Poole et~al.(2019)Poole, Ozair, van~den Oord, Alemi, and
  Tucker]{Poole2019}
Ben Poole, Sherjil Ozair, Aaron van~den Oord, Alexander~A. Alemi, and George
  Tucker.
\newblock {On variational bounds of mutual information}.
\newblock \emph{arXiv}, 2019.
\newblock ISSN 23318422.

\bibitem[Ranganath et~al.(2016)Ranganath, Tran, and Blei]{Ranganath2016}
Rajesh Ranganath, Dustin Tran, and David~M. Blei.
\newblock {Hierarchical Variational Models}.
\newblock \emph{ICML}, 33:\penalty0 1--9, 2016.

\bibitem[Salimans et~al.(2015)Salimans, Kingma, and Welling]{Salimans2015}
Tim Salimans, Diederik~P. Kingma, and Max Welling.
\newblock {Markov Chain Monte Carlo and Variational Inference: Bridging the
  Gap}.
\newblock \emph{Proceedings of the 32nd International Conference on Machine
  Learning}, pages 1218--1226, 2015.
\newblock URL \url{http://arxiv.org/abs/1410.6460}.

\bibitem[Sohl-Dickstein et~al.(2015)Sohl-Dickstein, Weiss, Maheswaranathan, and
  Ganguli]{sohl2015deep}
Jascha Sohl-Dickstein, Eric Weiss, Niru Maheswaranathan, and Surya Ganguli.
\newblock Deep unsupervised learning using nonequilibrium thermodynamics.
\newblock In \emph{International Conference on Machine Learning}, pages
  2256--2265. PMLR, 2015.

\bibitem[Stam(1959)]{Stam1959}
A.~J. Stam.
\newblock {Some inequalities satisfied by the quantities of information of
  Fisher and Shannon}.
\newblock \emph{Information and Control}, 2\penalty0 (2):\penalty0 101--112,
  1959.
\newblock ISSN 00199958.
\newblock \doi{10.1016/S0019-9958(59)90348-1}.

\bibitem[Stein and Shakarchi(2011)]{stein2011fourier}
Elias~M Stein and Rami Shakarchi.
\newblock \emph{Fourier analysis: an introduction}, volume~1.
\newblock Princeton University Press, 2011.

\bibitem[Virtanen et~al.(2020)Virtanen, Gommers, Oliphant, Haberland, Reddy,
  Cournapeau, Burovski, Peterson, Weckesser, Bright, {van der Walt}, Brett,
  Wilson, Millman, Mayorov, Nelson, Jones, Kern, Larson, Carey, Polat, Feng,
  Moore, {VanderPlas}, Laxalde, Perktold, Cimrman, Henriksen, Quintero, Harris,
  Archibald, Ribeiro, Pedregosa, {van Mulbregt}, and {SciPy 1.0
  Contributors}]{scipy}
Pauli Virtanen, Ralf Gommers, Travis~E. Oliphant, Matt Haberland, Tyler Reddy,
  David Cournapeau, Evgeni Burovski, Pearu Peterson, Warren Weckesser, Jonathan
  Bright, St{\'e}fan~J. {van der Walt}, Matthew Brett, Joshua Wilson, K.~Jarrod
  Millman, Nikolay Mayorov, Andrew R.~J. Nelson, Eric Jones, Robert Kern, Eric
  Larson, C~J Carey, {\.I}lhan Polat, Yu~Feng, Eric~W. Moore, Jake
  {VanderPlas}, Denis Laxalde, Josef Perktold, Robert Cimrman, Ian Henriksen,
  E.~A. Quintero, Charles~R. Harris, Anne~M. Archibald, Ant{\^o}nio~H. Ribeiro,
  Fabian Pedregosa, Paul {van Mulbregt}, and {SciPy 1.0 Contributors}.
\newblock {{SciPy} 1.0: Fundamental Algorithms for Scientific Computing in
  Python}.
\newblock \emph{Nature Methods}, 17:\penalty0 261--272, 2020.
\newblock \doi{10.1038/s41592-019-0686-2}.

\bibitem[Wainwright and Jordan(2008)]{Wainwright2008}
Martin~J. Wainwright and Michael~I. Jordan.
\newblock {Graphical Models, Exponential Families, and Variational Inference}.
\newblock \emph{Foundations and Trends{\textregistered} in Machine Learning},
  1\penalty0 (1–2):\penalty0 1--305, 2008.
\newblock ISSN 1935-8237.
\newblock \doi{10.1561/2200000001}.

\bibitem[Wei and Stocker(2016)]{Wei2016}
Xue-Xin Wei and Alan~A. Stocker.
\newblock {Mutual Information, Fisher Information, and Efficient Coding}.
\newblock \emph{Neural computation}, 28\penalty0 (2), 2016.
\newblock \doi{10.1162/NECO_a_0084}.

\bibitem[Wolf et~al.(2016)Wolf, Karl, and van~der Smagt]{wolf2016variational}
Christopher Wolf, Maximilian Karl, and Patrick van~der Smagt.
\newblock Variational inference with hamiltonian monte carlo.
\newblock \emph{arXiv preprint arXiv:1609.08203}, 2016.

\bibitem[Yin and Zhou(2018)]{Yin2018}
Mingzhang Yin and Mingyuan Zhou.
\newblock {Semi-Implicit Variational Inference}.
\newblock \emph{International Conference on Machine Learning}, 35, 2018.

\bibitem[Zhang et~al.(2019)Zhang, Butepage, Kjellstrom, and Mandt]{Zhang2019}
Cheng Zhang, Judith Butepage, Hedvig Kjellstrom, and Stephan Mandt.
\newblock {Advances in Variational Inference}.
\newblock \emph{IEEE Transactions on Pattern Analysis and Machine
  Intelligence}, 41\penalty0 (8):\penalty0 2008--2026, 2019.
\newblock ISSN 19393539.
\newblock \doi{10.1109/TPAMI.2018.2889774}.

\bibitem[Zhang et~al.(2021)Zhang, Carpenter, Gelman, and Vehtari]{Zhang2021}
Lu~Zhang, Bob Carpenter, Andrew Gelman, and Aki Vehtari.
\newblock {Pathfinder: Parallel quasi-Newton variational inference}.
\newblock \emph{arXiv}, 2021.
\newblock URL \url{http://arxiv.org/abs/2108.03782}.

\bibitem[Zobay(2014)]{Zobay2014}
O.~Zobay.
\newblock {Variational Bayesian inference with Gaussian-mixture
  approximations}.
\newblock \emph{Electronic Journal of Statistics}, 8\penalty0 (1):\penalty0
  355--389, 2014.
\newblock ISSN 19357524.
\newblock \doi{10.1214/14-EJS887}.

\end{thebibliography}

\null
\onecolumn
\vfill
\pagebreak

\setcounter{section}{0}
\renewcommand{\thesection}{\Alph{section}}
\setcounter{equation}{0}
\renewcommand{\theequation}{\Alph{section}.\arabic{equation}}
\setcounter{figure}{0}
\renewcommand{\thefigure}{\Alph{section}.\arabic{figure}}
\setcounter{table}{0}
\renewcommand{\thetable}{\Alph{section}.\arabic{table}}

\section{Proofs and Derivations}

Throughout, we assume that $\theta$ forms a minimal statistical manifold \citep{Amari2016}, so that the degrees of freedom of $\q$ match the dimensionality of $\theta$, and whenever $\q(\x;\theta_i) = \q(\x;\theta_j)$ for all $\x$, it must be that $\theta_i = \theta_j$.

Recall that in the main text, we defined the following objective:
\begin{equation}\tag{(\ref{eqn:weighted_optim}) restated}
    \L(\psi,\lambda) \equiv \MI[\x;\theta] - \lambda \E_{\psi(\theta)}\left[\KL(\q(\x;\theta)||\p^*(\x))\right] \, ,
\end{equation}
where $\lambda\in[1,\infty)$ is a hyper-parameter, and $\psi$ is a probability density on $\theta$. We also introduced an \textbf{approximate objective} in which $\MI[\x;\theta]$ is replaced with
\begin{equation}\tag{(\ref{eqn:mi_stams}) restated}
    \MI_\FIM[\x;\theta] \equiv \H[\theta] - \frac{1}{2}\E_{\psi(\theta)}\left[\log\left|2\pi e \FIM(\theta)^{-1}\right|\right] \, .
\end{equation}
This approximate objective is
\begin{equation}\tag{(\ref{eqn:weighted_optim_f}) restated}
    \L_\FIM(\psi,\lambda) = \H[\theta] + \E_{\psi(\theta)}\left[\frac{1}{2}\log|\FIM(\theta)| - \lambda \KL(\q(\x;\theta)||\p^*(\x))\right] \, ,
\end{equation}
and it is maximized for a given $\lambda$ by
\begin{align*}
    \psi(\theta) &= \frac{1}{Z}\exp\left(\frac{1}{2}\log|\FIM(\theta)| - \lambda \KL(\q(\x;\theta)||\p(\x))\right) \tag{(\ref{eqn:log_psi_stams}) restated}\\
    \text{where} \qquad Z &= \int_\theta \exp\left(\frac{1}{2}\log|\FIM(\theta)| - \lambda \KL(\q(\x;\theta)||\p(\x))\right) \rd\theta \, .
\end{align*}

\subsection{Characterizing the Pareto Front}\label{app:pareto}

Let us begin with a set of results regarding the shape of the Pareto front that connects VI to Sampling in Figure \ref{fig:mi_kl_space}.
\begin{lemma}\label{lem:concave}
    $\L(\psi,\lambda)$ is concave in $\psi$, i.e. $\L(\omega\psi_1 + (1-\omega)\psi_2, \lambda) \geq \omega\L(\psi_1,\lambda)+(1-\omega)\L(\psi_2,\lambda)$ for $0 \leq \omega \leq 1$. Further, $\L_\FIM(\psi,\lambda)$ is \emph{strictly} concave in $\psi$.
\end{lemma}
\paragraph{Proof:} The proof for $\L$ follows from the fact that $\E_{\psi(\theta)}\left[\KL(\q(\x;\theta)||\p^*(\x))\right]$ is \emph{linear} in $\psi$, and $\MI[\x;\theta]$ is known to be concave in the marginal distribution of either variable \citep{Braverman2011}. The proof for $\L_\FIM$ is similar: the $\E_{\psi(\theta)}\left[\frac{1}{2}\log|\FIM(\theta)|\right]$ term is linear in $\psi$, and $\H[\theta]$ is strictly concave in $\psi$. This can be seen, for instance, by taking the second variational derivative of $\H[\theta]$ with respect to $\psi$:
\begin{align*}
    \nabla_\psi^2 \H[\theta] \big\rvert_{\theta_i\theta_j} &= \nabla_\psi\left( \nabla_\psi\H[\theta] \big\rvert_{\theta_i}\right)\big\rvert_{\theta_j} \\
        &= \nabla_\psi\left( -\nabla_\psi\int_\theta \psi(\theta)\log\psi(\theta)\rd\theta \big\rvert_{\theta_i}\right)\big\rvert_{\theta_j} \\
        &= \nabla_\psi\left( -1 - \log\psi(\theta_i) \right)\big\rvert_{\theta_j} \\
        &= \begin{cases}
            -\frac{1}{\psi(\theta_i)} &\text{if $\theta_i=\theta_j$} \\
            0 &\text{otherwise} \, .
        \end{cases}
\end{align*}
Since $\psi(\theta)\geq 0$ everywhere, this implies that the curvature of $\H[\theta]$ is strictly negative at all values of $\theta$. \qed


\begin{lemma}\label{lem:pareto_slope} Let $\MI^*(\lambda)$ and $\E[\KL]^*(\lambda)$ denote the values of Mutual Information and Expected KL achieved by optima of $\L$ for a given $\lambda$. Then, $\lambda$ defines the slope of the Pareto front:
\begin{align*}
    \lambda = \frac{\rd \MI^*/\rd \lambda}{\rd \E[\KL]^* / \rd \lambda} \, .
\end{align*}
Or, in the case of $\L_\FIM$, $\lambda$ similarly defines the slope of 
\begin{align*}
    \lambda = \frac{\rd \MI_\FIM^*/\rd \lambda}{\rd \E[\KL]^* / \rd \lambda} \, ,
\end{align*}
with $\MI_\FIM$ in place of $\MI$.
\end{lemma}
\paragraph{Proof:} This follows from viewing $\L$ as the Lagrangian of a constrained optimization problem, with $\lambda$ as a Lagrange multiplier. The same argument applies to both $\L$ and $\MI$ as to $\L_\FIM$ and $\MI_\FIM$, so we will just give the proof for one. Consider the constrained optimization problem of maximizing $\MI$ (or $\MI_\FIM$) subject to the constraint that $\E[\KL(\q||\p)]=C$. The Lagrangian for this problem is identical to (\ref{eqn:weighted_optim}), but with $C$ added:
\begin{align*}
    \L(\psi,\lambda) \equiv \MI[\x;\theta] - \lambda \left(\E_{\psi(\theta)}\left[\KL(\q(\x;\theta)||\p^*(\x))\right] - C\right)
\end{align*}
Optimizing with respect to $\psi$, this is a concave maximization problem with a linear constraint. A well-known property of such problems is that, at the solution, the Lagrange multiplier ($\lambda$) is equal to the change in the objective ($\MI^*$) per change in the constraint ($C$), or $\lambda = \frac{\rd \MI^*}{\rd C}$. Since $C$ is the constrained value of $\E[\KL(\q||\p)]$, we also immediately have $\frac{\rd \E[\KL]^*}{\rd C} = 1$. This implies that
\begin{align*}
    \lambda = \frac{\rd \MI_\FIM^*/\rd C}{\rd \E[\KL]^* / \rd C} \, .
\end{align*}
So far, we have treated $\lambda$ as a function of $C$, but for all values of $\lambda$ that correspond to a unique $C$, we can invert this relationship and treat $C$ as a function of $\lambda$. Then, assuming $\frac{\rd C}{\rd \lambda} \neq 0$ for all $1 \leq \lambda < \infty$ that we are interested in, we have
\begin{align*}
    \lambda &= \frac{\rd \MI_\FIM^*/\rd C \times \rd C / \rd \lambda}{\rd \E[\KL]^* / \rd C \times \rd C / \rd \lambda} = \frac{\rd \MI_\FIM^*/\rd \lambda}{\rd \E[\KL]^* / \rd \lambda} \, .
\end{align*}
Again using the fact that $C=\E[\KL]^*$ by construction, the condition that $\frac{\rd C}{\rd\lambda}\neq 0$ is equivalent to $\frac{\rd\E[\KL]^*}{\rd\lambda} \neq 0$. In other words, as long as changing $\lambda$ has some effect on $\E[\KL]^*$, the combined effect on $\MI^*$ and $\E[\KL]^*$ will be such that $\lambda=\frac{\rd \MI^*}{\rd \E[\KL]^*}$. \qed

\subsection{Sampling-like behavior of our method}\label{app:sampling}

Recall our definition of sampling:
\begin{definition}[Sampling]
A stochastic mixture, defined by the component family $\q(\x;\theta)$ and mixing distribution $\psi(\theta)$, is considered to be ``sampling'' if (i) it is \textbf{unbiased} in the limit of infinitely many components, i.e. $\m(\x) \rightarrow \p(\x)$; and, (ii) it consists of \textbf{non-overlapping components}. That is, for small values of of $0 < \epsilon \ll 1$, wherever $\q(\x;\theta_i) > \epsilon$, with high probability $\q(\x;\theta_j) < \epsilon$, for all pairs $\theta_i, \theta_j$ drawn independently from $\psi(\theta)$.
\end{definition}

We will assume throughout this section that $\q$ is a location-scale family, and in particular Gaussian for Lemma \ref{lem:sampling_approximate_solution}, but it may be fruitful for future work to consider other families of mixture components.


\begin{lemma}\label{lem:sampling_solution}
Sampling is an optimum of the original objective, $\L$, when $\lambda = 1$.
\end{lemma}
\paragraph{Proof:} When $\lambda = 1$, $\L$ simplifies back to $\KL(\m||\p)$. Any \textbf{unbiased} mixture is a minimum of $\KL(\m||\p)$. \qed

Note, however, that this does not imply sampling is the unique optimum. In general there may be other unbiased mixing distributions $\psi(\theta)$ such that $\m(\x)=\p(\x)$. For instance, if $\q$ is Gaussian and $\p(\x)$ is itself a finite mixture of Gaussians, then $\psi(\theta)$ could concentrate on exactly those modes in $\p$. In any case where there two such unbiased $\psi$s, there are in fact infinitely many unbiased, since any mixture of them, $\alpha\psi_1(\theta) + (1-\alpha)\psi_2(\theta)$, will also be unbiased. Among all unbiased mixtures, sampling may in some sense be the worst choice -- we conjecture that it has the highest variance of all unbiased mixtures.

\begin{lemma}\label{lem:sampling_approximate_solution}
When $\q$ is Gaussian and $\lambda = 1$, the optimal $\psi$ that maximizes the approximate objective $\L_\FIM$ is both \textbf{unbiased} and has \textbf{non-overlapping components}.
\end{lemma}
In other words, Lemma \ref{lem:sampling_approximate_solution} states that the solution to the approximate objective $\L_\FIM$ ``looks like'' sampling when $\lambda=1$, in the sense of Definition \ref{def:sampling}.
\paragraph{Proof:} Without loss of generality, let us assume that $\theta$ is already parameterized in terms of its location and scale, $[\bmu, \bsigma]$, where $\bmu$ determines the mean of $\q$ and $\bsigma$ determines its covariance. Then, the Fisher Information Matrix is a block-diagonal matrix:\footnote{\url{https://en.wikipedia.org/wiki/Fisher\_information\#Multivariate_normal_distribution}}
\begin{align*}
    \FIM(\theta) = \begin{bmatrix}\FIM(\bmu) & 0 \\ 0 & \FIM(\bsigma) \end{bmatrix}
\end{align*}
where
\begin{align*}
    \FIM(\bmu) &= \Lambda \\
    \FIM(\bsigma)_{ij} &= \frac{1}{2}\text{Tr}\left(\Lambda\frac{\partial \Sigma}{\partial \bsigma_i}\Lambda\frac{\partial \Sigma}{\partial \bsigma_j}\right) \, .
\end{align*}
$\Lambda$ and $\Sigma$ are the precision matrix and covariance matrix of $\q$, respectively. Both $\Lambda$ and $\Sigma$ are functions of the parameters $\bsigma$ but not of $\bmu$. To simplify further, let us assume that the covariance of $\q$ is diagonal, and that $\bsigma_i$ is the log standard deviation of the $i$th dimension of $\x$:
\begin{align*}
    \Sigma(\bsigma)_{ij} = \begin{cases}
        e^{2\bsigma_i} &\text{if $i=j$} \\
        0 &\text{otherwise}
    \end{cases}
\end{align*}
We emphasize that this simplification is for notational convenience only, and other parameterizations of $\Sigma(\bsigma)$ are permissible. With this assumption, $\FIM(\bsigma)$ becomes the identity matrix, and the log determinant of $\FIM(\theta)$ becomes simply
\begin{align*}
    \log|\FIM(\theta)| = \log|\Lambda| \, .
\end{align*}
So, for Gaussian $\q$, the expression for $\psi$ becomes
\begin{align*}
    \log\psi(\theta) = \log\psi(\bmu,\bsigma) = \frac{1}{2}\log|\Lambda(\bsigma)| - \lambda\KL(\q(\x;\bmu,\bsigma)||\p(\x)) \, .
\end{align*}

Next, we will split $\KL(\q||\p)$ into separate entropy and cross-entropy terms:
\begin{align*}
    \KL(\q||\p) &= \E_{\q(\x;\theta)}\left[\log \q(\x;\theta) \right] - \E_{\q(\x;\theta)}\left[\log \p(\x) \right] \\
        &= -\H[\q] + \CE[\q||\p] \, .
\end{align*}
And note that when $\q$ is Gaussian, its entropy is given by
\begin{align*}
    \H[\q] = \frac{1}{2}\log|2\pi e \Sigma| = \frac{1}{2}\log|\Sigma| + \text{constants} \, .
\end{align*}
Taking $\lambda=1$ and using the fact that $\log|\Sigma|=-\log|\Sigma^{-1}|=-\log|\Lambda|$ and combining the above three equations, the $\H[\q]$ and $\log|\FIM(\bmu)|$ terms cancel in $\psi$ and we are left -- up to additive constants -- with
\begin{equation}\label{eqn:log_psi_equals_ce}
    \log\psi(\theta) = -\CE[\q||\p] = \E_{\q(\x;\bmu,\bsigma)}\left[\log \p(\x) \right] \, .
\end{equation}
To summarize, equation (\ref{eqn:log_psi_equals_ce}) says that, using Gaussian components and letting $\lambda\rightarrow 1$, our method, derived from the $\MI_\FIM$ approximation to $\MI$, selects components simply according to the \emph{cross entropy} between $\q(\x;\theta)$ and $\p(\x)$.

Note that (\ref{eqn:log_psi_equals_ce}) is not a proper distribution over $\theta$. To see this, consider any sufficiently narrow component such that $\q$ behaves like a Dirac delta, or $\E_{\q(\x;\bmu,\bsigma)}[\log\p(\x)] \approx \log\p(\bmu)$. Wherever this holds for some $\bsigma$, it will additionally hold for all \emph{narrower} components at the same $\bmu$.\footnote{There is an implicit assumption here that $\log\p(\x)$ is almost everywhere smooth, so that there is some small enough scale at which $\p(\x)$ appears locally linear under $\q$.} Therefore, below a particular scale where $\q$ behaves like a Dirac delta, (\ref{eqn:log_psi_equals_ce}) places uniform mass on the infinitely many $\q$s that are at least as narrow. This effect is visible in the top-right panel of Figure \ref{fig:mi_kl_space}. Also note that $\psi$ is only improper for $\lambda=1$; for all other $\lambda > 1$, a $(\lambda-1)\H[\q]$ term remains, and $\psi$ cannot place arbitrarily much mass on arbitrarily narrow components.

Despite its impropriety, we are free to draw samples of $\theta$ from this improper $\psi$ when $\lambda=1$ \citep{Besag1995,Hobert1996}. We will then find that with probability approaching $1$ we only ever see components that ``look like'' Dirac-deltas. This phenomenon is seen empirically in all of our experiments where we set $\lambda=1$ and run HMC dynamics drawing $\theta \sim \psi(\theta)$. Since components will become arbitrarily narrow, we have the \textbf{non-overlapping components} property required by our definition of sampling.

Consider decomposing $\psi(\theta)$ into $\psi(\bsigma)\psi(\bmu|\bsigma)$. The previous paragraph establishes that the marginal distribution $\psi(\bsigma)$ will allocate effectively all samples to parts of $\theta$-space where components behave like Dirac deltas. This implies
\begin{align*}
    \log\psi(\bmu|\bsigma = \text{narrow}) &= \E_{\q(\x;\bmu,\bsigma)}\left[\log \p(\x) \right] \\
        &= \log \p(\bmu) \, .
\end{align*}
Hence, $\m(\x)$ will be a mixture of Dirac-delta-like components, each of which is chosen in proportion to the true probability of its mean, $\p(\bmu)$. This means that $\m(\x)$ will be \textbf{unbiased}. \qed

\begin{theorem}[Improve on sampling]
If a mixture is sampling as in Definition \ref{def:sampling}, then $\frac{\rd}{\rd\lambda}\text{KL bias}=0$ and $\frac{\rd}{\rd\lambda}\text{KL variance} < 0$. Thus, $\frac{\rd}{\rd\lambda}\text{KL error}<0$.
\end{theorem}
\paragraph{Proof:} Our approach will be to calculate the variational derivatives of KL bias and KL error with respect to $\psi$, then take the inner product (directional derivative) with the change in $\psi$ per change in $\lambda$.

First, we need the sensitivy of $\psi$ to changes in $\lambda$. Recall that the closed-form solution for $\psi$ we get from solving $\L_\FIM$ is
\begin{align*}
    \log\psi(\theta) = \frac{1}{2}\log|\FIM(\theta)| - \lambda\KL(\q(\x;\theta)||\p(\x)) - \log Z(\lambda) \, .
\end{align*}
The sensitivity of $\log\psi$ to $\lambda$ is
\begin{align*}
\frac{\rd}{\rd \lambda}\log\psi(\theta) &= -\KL(\q||\p) + \frac{1}{Z}\int_{\theta'}e^{\frac{1}{2}\log|\FIM(\theta)| - \lambda \KL(\q||\p)} \KL(\q||\p) \rd\theta'  \\
    &= \E_\psi[\KL(\q||\p)] - \KL(\q||\p) \, .
\end{align*}
Converting from $\log\psi$ to $\psi$, we get
\begin{equation}\label{eqn:dpsi_dlambda}
    \frac{\rd}{\rd \lambda}\psi(\theta) = \psi(\theta)\left(\E_\psi[\KL(\q||\p)] - \KL(\q||\p)\right)
\end{equation}

Recall that we defined $\text{KL bias}=\KL(\m||\p)$ and $\text{KL variance}=\E[\KL(\m_T||\m)]$. The variational derivative of Bias with respect to $\psi$, evaluated at $\rtheta$ is
\begin{align}
\nabla_\psi \KL(\m||\p) =& \nabla_\psi \int_\x \left(\E_\psi[\q(\x;\theta)]\right)\log\frac{\left(\E_\psi[\q(\x;\theta)]\right)}{\p(\x)} \rd\x \nonumber \\
    =& \int_\x \left(\m(\x)\frac{\p(\x)}{\m(\x)}\frac{\q(\x;\rtheta)}{\p(\x)} + \q(\x;\rtheta)\log\frac{\m(\x)}{\p(\x)} \right)\rd\x \nonumber \\
    =& 1 + \E_{\q(\x;\rtheta)}\left[\log\frac{\m(\x)}{\p(\x)}\right] \, . \label{eqn:dbias_dpsi}
\end{align}
To get the sensitivity of Bias to $\lambda$ we will take the inner-product of (\ref{eqn:dpsi_dlambda}) with (\ref{eqn:dbias_dpsi}). This is
\begin{align*}
\frac{\rd}{\rd \lambda} \text{KL bias} &= \left\langle\frac{\rd \text{KL bias}}{\rd \psi}, \frac{\rd \psi}{\rd\lambda}\right\rangle & \\
    &= \int_\rtheta \left(1 + \E_{\q(\x;\rtheta)}\left[\log\frac{\m(\x)}{\p(\x)}\right] \right) \psi(\rtheta)\left(\E_\psi[\KL(\q||\p)] - \KL({\color{red} \q}||\p)\right) \rd\rtheta & \\
    &= \int_\rtheta (1 + 0) \psi(\rtheta)\left(\E_\psi[\KL(\q||\p)] - \KL({\color{red}\q}||\p)\right) \rd\rtheta & \text{(\textbf{unbiased})} \\
    &= \E_\psi[\KL(\q||\p)] - \E_\psi[\KL(\q||\p)] & \\
    &= 0 \, . &
\end{align*}
So, we can conclude that in the sampling limit, small changes in $\lambda$ have no effect on Bias. Geometrically, this tells us the Pareto Front is tangent to the y=x line in that limit, as illustrated in Figure \ref{fig:mi_kl_space}.

Next we will consider the variational derivative of the Variance component of KL error with respect to $\psi$, where
\begin{align*}
    \text{KL variance} &\equiv \E_{1..T}[\KL(\m_T||\m)] \\
        &= \E_{1..T}\left[\int_\x\left(\frac{1}{T}\sum_{t=1}^T\q(\x;\theta_t)\right)\log\frac{\left(\frac{1}{T}\sum_{j=1}^T\q(\x;\theta_j)\right)}{\m(\x)}\rd\x\right]
\end{align*}
using the shorthand $\E_{1..T}[\ldots]$ to denote an expectation over independent draws of $\lbrace{\theta_t}\rbrace \sim \psi(\theta)$. We will apply the assumption of \textbf{non-overlapping components} to simplify $\KL(\m_T||\m)$. Let $\int_{\x\in\q_t}\ldots\rd\x$ denote an integral over just the region of $\x$-space where $\q(\x;\theta_t)>\epsilon$ for some small $\epsilon$. By assumption, these regions are disjoint for all pairs of $\theta$s, with high probability. Splitting the integral into $T$ separate regions and rearranging terms inside the $\log$, we have
\begin{align*}
    \text{KL variance} &\approx \frac{1}{T}\sum_{t=1}^T\E_{1..T}\left[\int_{\x\in\q_t} \q(\x;\theta_t) \log\left(\frac{\q(\x;\theta_t)}{\m(\x)}\left(\frac{1}{T} + \frac{1}{T}\sum_{j\neq t}\frac{\q(\x;\theta_j)}{\q(\x;\theta_t)}\right) \right) \rd\x\right] + \mathcal{O}(\epsilon) \\
        &= \frac{1}{T}\sum_{t=1}^T\E_{1..T}\left[\int_{\x\in\q_t} \q(\x;\theta_t) \left(\log\frac{\q(\x;\theta_t)}{\m(\x)} + \log \left(\frac{1}{T} + \frac{1}{T}\sum_{j\neq t}\frac{\q(\x;\theta_j)}{\q(\x;\theta_t)}\right) \right) \rd\x\right] + \mathcal{O}(\epsilon) \\
        &= \E_\psi[\KL(\q||\m)] + \frac{1}{T}\sum_{t=1}^T\E_{1..T}\left[\int_{\x\in\q_t} \q(\x;\theta_t) \log \left(\frac{1}{T} + \frac{1}{T}\sum_{j\neq t}\frac{\q(\x;\theta_j)}{\q(\x;\theta_t)}\right) \rd\x\right] + \mathcal{O}(\epsilon) \\
\end{align*}
From here, we can get an upper-bound on Variance by noting that $\frac{\q(\x;\theta_j)}{\q(\x;\theta_t)}\leq 1$ by the non-overlapping assumption. Since there are $T-1$ of these in the sum, $\log\left(\frac{1}{T} + \frac{1}{T}\sum_{j\neq t}\frac{\q(\x;\theta_j)}{\q(\x;\theta_t)}\right) \leq \left(\frac{1}{T} + \frac{T-1}{T}\right) = 0$, and so
\begin{align*}
    \text{KL variance} \leq \E_\psi[\KL(\q||\m)] + \mathcal{O}(\epsilon) \, .
\end{align*}
Note that this bound used the assumption of non-overlapping components, and is therefore only applicable for small $\lambda$ and moderate values of $T$. Since we are interested in showing that $\frac{\rd}{\rd \lambda}\text{KL variance} < 0$ in the sampling limit, showing that the \emph{upper bound} on variance decreases with $\lambda$ will suffice. Using this upper-bound, we get the following variational derivative of Variance with respect to $\psi$ at each value of $\rtheta$:
\begin{align*}
    \nabla_\psi\text{KL variance}\big\rvert_{\rtheta} &\approx \nabla_\psi\left. \int_\theta \psi(\theta) \int_\x \q(\x;\theta) \log\frac{\q(\x;\theta)}{\m(\x)}\rd\x\rd\theta \right\rvert_{\rtheta} \\
        &= -\int_\theta \psi(\theta) \int_\x \q(\x;\theta)\frac{\m(\x)}{\q(\x;\theta)}\frac{\q(\x;\theta)}{\m(\x)^2}\q(\x;\rtheta)\rd\x\rd\theta + \int_\x \q(\x;\rtheta) \log\frac{\q(\x;\rtheta)}{\m(\x)}\rd\x \\
        &= -1 + \KL(\q(\x;\rtheta)||\m(\x)) \, .
\end{align*}
Taking the inner product with $\frac{\rd}{\rd\lambda}\psi$, and applying the \textbf{unbiased} assumption,
\begin{align*}
\frac{\rd}{\rd \lambda} \text{KL variance} &= \left\langle\frac{\rd \text{KL variance}}{\rd \psi}, \frac{\rd \psi}{\rd\lambda}\right\rangle & \\
    &= \int_\rtheta \left(-1 + \KL({\color{red}\q}||\m) \right) \psi(\rtheta) \left(\E_\psi[\KL(\q||\p)] - \KL({\color{red}\q}||\p)\right) \rd\rtheta & \\
    &= \int_\rtheta \left(-1 + \KL({\color{red}\q}||\p) \right) \psi(\rtheta) \left(\E_\psi[\KL(\q||\p)] - \KL({\color{red}\q}||\p)\right) \rd\rtheta & \text{(\textbf{unbiased})}\\
    &= -\E_{\psi(\rtheta)}\left[\left(\KL({\color{red}\q}||\p) - \E_\psi[\KL(\q||\p)]\right)\KL({\color{red}\q}||\p)\right] & \\
    &= -\text{var}\left(\KL(\q||\p)\right) \, .
\end{align*}
In other words, this says that the change in the (upper bound on) ``Variance,'' defined as $\E_{1..T}\KL(\m_T||\m)]$, is \emph{negative}, with magnitude given by the variance of the values taken by $\KL(\q||\p)$ across all $\theta$.

To summarize, we have shown that, in the sampling limit, where $\lambda=1$, we have $\frac{\rd}{\rd\lambda}\text{KL bias} = 0$ and $\frac{\rd}{\rd\lambda}\text{KL variance} \leq 0$, which proves the lemma. \qed

\subsection{VI-like behavior of our method}\label{app:vi}

\begin{definition}[VI limit]\label{def:vi_limit} We model the large $\lambda$ limit of our method using a Laplace approximation around the optimal $\thetastar=\argmin_\theta \KL(\q(\x;\theta)||\p(\x))$:
\begin{equation}\label{eqn:define_vi}
\begin{split}
    \psi(\theta) &\approx \N(\theta; \thetastar, \Sigmastar) \\
    \text{where} \qquad \Sigmastar^{-1} &= \lambda \nabla^2_\theta \KL(\q(\x;\theta)||\p(\x))\big\rvert_{\thetastar} \, .
\end{split}
\end{equation}
\end{definition}
In other words, we approximate $\psi$ by a normal distribution whose mean is $\thetastar$ and whose precision is set by the curvature of $\KL(\q(\x;\theta)||\p(\x))$ and scales with $\lambda$. We will assume, for the purposes of proofs related to the VI limit, that there is a single optimal $\thetastar$.


\begin{theorem}[Improve on VI]
Assume that $\p(\x)$ is heaver-tailed than $\q(\x;\thetastar)$. Then, there exists some $T_0>1$ such that for all $T\geq T_0$, $\frac{\rd}{\rd\lambda}\text{KL error}>0$, when $\lambda$ is sufficiently large.
\end{theorem}
\paragraph{Proof:} As $\lambda$ grows, the Laplace approximation in (\ref{eqn:define_vi}) becomes increasingly narrow. This allows us to approximate expectations under $\psi$ using a second order Taylor approximation to the integrand. The general rule for multivariate Gaussians is
\begin{align*}
    \E_{\N(\mathbf{y};\mu,\Sigma)}[f(\mathbf{y})] \approx f(\mu) + \frac{1}{2}\text{Tr}\left(\Sigma \; \nabla^2_\mathbf{y}f\right)\big\rvert_\mu
\end{align*}
Recall that we defined KL error as $\E_{1..T}[\KL(\m_T(\x)||\p(\x))]$. Approximating each $\psi(\theta_t)$ as a multivariate Gaussian, their product is also a multivariate Gaussian whose collective covariance is block-diagonal\footnote{This assumes the $T$ components are statistically independent draws from $\psi(\theta)$. The approach outlined here could be generalized to include correlation between $\theta$s in the off-block-diagonals to model variance of an autocorrelated chain of $\theta$ values.} containing $T$ copies of $\Sigmastar$ from (\ref{eqn:define_vi}), and whose collective mean is $\thetastar$ for each component. At this mean value where all $T$ components' parameters are equal to $\thetastar$, $\m_T(\x)$ becomes $\q(\x;\thetastar)$. Hence, applying the Taylor series approximation to KL error, the $f(\mu)$ term is just $\KL(\q(\x;\thetastar)||\p(\x))$. The second term is
\begin{align*}
    \frac{1}{2}\text{Tr}\left(
        \begin{bmatrix}
            \Sigmastar &  &  & 0  \\
             & \Sigmastar & & &  \\
             &  & \ddots & \\
             0 &  &  & \Sigmastar
        \end{bmatrix}
        \nabla^2_{\theta_1,\ldots,\theta_T}\KL(\m_T||\p)
    \right) \, .
\end{align*}
First, note that the zeros in the off-block-diagonal terms on the left mean that we can ignore interactions between $\theta$s across different mixture components in the Hessian term on the right. Second, there is $T-$fold symmetry between all components. So, this simplifies to
\begin{align*}
    \frac{T}{2}\text{Tr}\left(\Sigmastar \; \nabla^2_{\theta_1}\KL(\m_T||\p) \right) = \frac{T}{2\lambda}\text{Tr}\left((\nabla^2_\theta\KL(\q||\p))^{-1} \; \nabla^2_{\theta_1}\KL(\m_T||\p) \right) \, .
\end{align*}

Next, since this Hessian is being evaluated around the point $\thetastar$, all of $\theta_2,\ldots,\theta_T$ are equal to $\thetastar$, and we can write the mixture as a function only of the component parameters we are varying in the Hessian. Call this mixture with $T-1$ components set to the variational solution $\m_T^*$, defined as
\begin{align*}
    \m_T^*(\x;\theta) = \frac{T-1}{T}\q(\x;\thetastar) + \frac{1}{T}\q(\x;\theta) \, .
\end{align*}

We will now calculate each of these Hessians. Note: in what follows we will use $\theta_i$ and $\theta_j$ to indicate the $i$ and $j$th indices of the vector $\theta$, whereas we had used $\theta_t$ to indicate one of $T$ vectors. For $\Sigmastar$, we need the second derivative (Hessian) of $\KL(\q||\p)$:
\begin{align}
    \frac{\partial^2}{\partial\theta_j\partial\theta_i} \KL(\q(\x;\theta)||\p(\x)) &= \frac{\partial^2}{\partial\theta_j\partial\theta_i} \int_\x\q(\x;\theta)\log\frac{\q(\x;\theta)}{\p(\x)}\rd\x \nonumber \\
        &= \frac{\partial}{\partial\theta_j} \int_\x\left[\left(\frac{\partial}{\partial\theta_i}\q(\x;\theta)\right)\left(1 +  \log\frac{\q(\x;\theta)}{\p(\x)} \right)\right]\rd\x \nonumber \\
        &= \int_\x\left[\left(\frac{\partial}{\partial\theta_i}\q(\x;\theta)\right)\left(\frac{\frac{\partial}{\partial\theta_j}\q(\x;\theta)}{\q(\x;\thetastar)} \right) + \left(\frac{\partial^2}{\partial\theta_i\partial\theta_j}\q(\x;\theta)\right)\left(1 +  \log\frac{\q(\x;\thetastar)}{\p(\x)} \right)\right]\rd\x \nonumber \\
        (*) &= \int_\x \frac{\left(\frac{\partial}{\partial\theta_i}\q(\x;\theta)\right)\left(\frac{\partial}{\partial\theta_j}\q(\x;\theta)\right)}{\q(\x;\thetastar)} \rd\x + \int_\x \left(\frac{\partial^2}{\partial\theta_i\partial\theta_j}\q(\x;\theta)\right)\log\frac{\q(\x;\thetastar)}{\p(\x)} \rd\x \nonumber \\
        &= \FIM(\thetastar) + M(\thetastar) \label{eqn:hess_kl_q_p} \,  .
\end{align}
In line $(*)$ we used the fact that $\int_\x \nabla^2_\theta \q(\x;\theta) \rd\x = \nabla^2_\theta \int_\x \q(\x;\theta) \rd\x = \nabla^2_\theta 1 = 0$. $\FIM$ is the Fisher Information Matrix, and we have defined $M(\theta) = \int_\x\left(\frac{\partial^2}{\partial\theta_i\partial\theta_j}\q(\x;\theta)\right)\log\frac{\q(\x;\theta)}{\p(\x)}\rd\x$.

Following a similar derivation, the Hessian of $\KL(\m_T^*(\x;\theta)||\p(\x))$ is
\begin{align}
    \frac{\partial^2}{\partial\theta_j\partial\theta_i} & \KL(\m_T^*(\x;\theta)||\p(\x)) = \frac{\partial^2}{\partial\theta_j\partial\theta_i} \int_\x\left(\frac{T-1}{T}\q(\x;\thetastar)+\frac{1}{T}\q(\x;\theta)\right)\log\frac{\left(\frac{T-1}{T}\q(\x;\thetastar)+\frac{1}{T}\q(\x;\theta)\right)}{\p(\x)}\rd\x \nonumber \\
        &= \frac{\partial}{\partial\theta_j} \int_\x\left[\frac{1}{T}\left(\frac{\partial}{\partial\theta_i}\q(\x;\theta)\right) + \frac{1}{T}\left(\frac{\partial}{\partial\theta_i}\q(\x;\theta)\right)\log\frac{\left(\frac{T-1}{T}\q(\x;\thetastar)+\frac{1}{T}\q(\x;\theta)\right)}{\p(\x)} \right]\rd\x \nonumber \\
        &= \frac{1}{T} \frac{\partial}{\partial\theta_j} \int_\x\left[\left(\frac{\partial}{\partial\theta_i}\q(\x;\theta)\right)\left(1 + \log\frac{\left(\frac{T-1}{T}\q(\x;\thetastar)+\frac{1}{T}\q(\x;\theta)\right)}{\p(\x)}\right) \right]\rd\x \nonumber \\
        &= \frac{1}{T} \int_\x\left[\left(\frac{\partial}{\partial\theta_i}\q(\x;\theta)\right)\left(\frac{\frac{1}{T}\frac{\partial}{\partial\theta_j}\q(\x;\theta)}{\m_T^*(\x;\theta)}\right) + \left(\frac{\partial^2}{\partial\theta_i\partial\theta_j}\q(\x;\theta)\right)\left(1 + \log\frac{\m_T^*(\x;\theta)}{\p(\x)}\right) \right]\rd\x \nonumber \\
   (**) &= \frac{1}{T^2}\int_\x\frac{\left(\frac{\partial}{\partial\theta_i}\q(\x;\theta)\right)\left(\frac{\partial}{\partial\theta_j}\q(\x;\theta)\right)}{\q(\x;\thetastar)}\rd\x + \frac{1}{T}\int_\x\left(\frac{\partial^2}{\partial\theta_i\partial\theta_j}\q(\x;\theta)\right)\log\frac{\q(\x;\thetastar)}{\p(\x)}\rd\x \nonumber \\
        &= \frac{1}{T^2}\FIM(\thetastar) + \frac{1}{T} M(\thetastar) \nonumber \\
        &= \frac{1}{T}\frac{\partial^2}{\partial\theta_j\partial\theta_i} \KL(\q(\x;\theta)||\p(\x)) + \FIM(\theta)\left(\frac{1-T}{T^2}\right) 
        \label{eqn:hess_kl_mt_p}
\end{align}
Here, in $(**)$, we additionally used the fact that $\m_T^*(\x;\thetastar)=\q(\x;\thetastar)$. We then wrote the final line in terms of the Hessian of $\KL(\q||\p)$ in (\ref{eqn:hess_kl_q_p}).

To summarize, near the variational limit we have that the KL error is approximately
\begin{align*}
    \KL(\q(\x;\thetastar)||\p(\x)) + \frac{T}{2\lambda}\text{Tr}((\underbrace{\nabla^2_\theta\KL(\q||\p)}_{(\ref{eqn:hess_kl_q_p})})^{-1} \; (\underbrace{\nabla^2_\theta\KL(\m_T^*||\p)}_{(\ref{eqn:hess_kl_mt_p})})) \, .
\end{align*}
Plugging in (\ref{eqn:hess_kl_q_p}) and (\ref{eqn:hess_kl_mt_p}), this is
\begin{align*}
    \text{KL error} &\approx \KL(\q(\x;\thetastar)||\p(\x)) + \frac{1}{2\lambda}\text{Tr}\left(\mathbf{I} + \frac{1-T}{T}(\FIM+M)^{-1} \FIM\right) \\
        &= \KL(\q(\x;\thetastar)||\p(\x)) + \frac{d}{2\lambda} - \frac{1}{2\lambda}\text{Tr}\left(\frac{T-1}{T}(\FIM+M)^{-1} \FIM\right)
\end{align*}
where $\mathbf{I}$ is the identity matrix.
Consider the case where $T=1$: the KL error simplifies to $\KL(\q(\x;\thetastar)||\p(\x)) + \frac{d}{2\lambda}$ where $d$ is the dimensionality of $\theta$. Therefore when $T=1$, KL error is only reduced by further increasing $\lambda$. This is an intuitive result: we cannot reduce bias compared to VI when using a single component, and any stochasticity only adds variance.

Now consider the case where $T \geq 2$. We are interested in cases where KL error \emph{increases} with $\lambda$ near the VI limit. This is equivalent to asking when the following inequality holds:
\begin{align*}
    \text{Tr}\left((\FIM+M)^{-1} \FIM\right) > \frac{T}{T-1}\overbrace{\text{Tr}(\mathbf{I})}^{d} \, ,
\end{align*}
Recall from (\ref{eqn:hess_kl_q_p}) that $\FIM+M$ is the Hessian of $\KL(\q(\x;\theta)||\p(\x))$, and note that the Fisher Information Matrix is equivalent to the Hessian with respect to $\theta$ of $\KL(\q(\x;\theta)||\q(\x;\thetastar))$, so we can rewrite this inequality as
\begin{align*}
    \text{Tr}\left((\nabla^2_\theta\KL(\q||\p))^{-1} \nabla^2_\theta\KL(\q||\q^*)\right) > \frac{T}{T-1}\text{Tr}(\mathbf{I}) \, .
\end{align*}
Assuming $\thetastar$ is a local minimum of $\KL(\q||\p)$ (which follows from the assumption that $\thetastar$ is the unique minimum), both of these are positive definite matrices encoding how sharply curved the $\KL(\q||\p)$ or $\KL(\q||\q^*)$ objectives are.

If $\p$ is in the same family as $\q$, then $\q*=\p$ and this inequality becomes $\text{Tr}\left(\mathbf{I}\right) > \frac{T}{T-1}\text{Tr}(\mathbf{I})$, which is false for all finite $T$ and approaches equality as $T\rightarrow\infty$. This again captures the intuitive idea that we cannot improve on VI by reducing $\lambda$ when the single-component $\q$ is already unbiased.

Conversely, we can view the ratio
\begin{align}
    \frac{\text{Tr}\left((\nabla^2_\theta\KL(\q||\p))^{-1} \nabla^2_\theta\KL(\q||\q^*)\right)}{\text{Tr}(\mathbf{I})} \label{eqn:ratio_hessian_match}
\end{align}
as an indication of \emph{how poorly matched} $\q(\x;\thetastar)$ is to $\p(\x)$, locally around the single-component VI solution. We \textbf{conjecture} that this ratio is always greater than $1$ whenever $\p(\x)$ is heavier-tailed than $\q(\x;\thetastar)$. Since $\frac{T}{T-1}$ approaches $1$ from above in the limit of large $T$, this implies that there will be some finite $T_0$ where $\frac{T_0}{T_0-1}$ is less than the ratio in (\ref{eqn:ratio_hessian_match}), and that such a $T_0$ will be reached sooner the worse $\q(\x;\thetastar)$ locally approximates $\p$. \qed

\vfill\pagebreak
\section{Additional Experiments}
\label{app:posteriordb}

\begin{figure*}[ht]
    \centering
    \includegraphics[width=\textwidth]{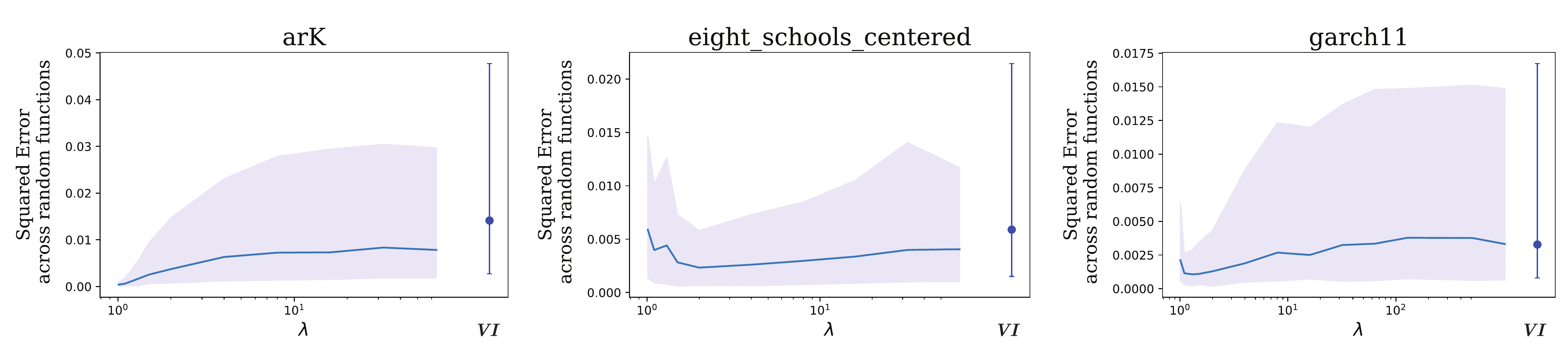}
    \caption{Further examples. To verify that the results in the main manuscript generalize other datasets, we ran our algorithm at multiple values of $\lambda$ on three problems from the posteriordb dataset \citep{posteriordb}. Rather than show convergence for one random $f(\x)$, as in the main text, we generated 200 random functions over the unconstrained parameter space of each model, using the Fourier synthesis method described in section \ref{app:numerical_details} below, with $\alpha=-1$. We then calculated the expectation $\E_{\m_T(\x)}[f(\x)]$ after subsampling mixtures of $T=100$ components, and plotted the distribution of squared error relative to ground truth expectations (based on very long runs of Stan's default NUTS implementation). Error bars reflect the combined effect of two sources of variability: one from the random choice of $f(\x)$ and one from the random subsampling of mixtures. Blue line is the median squared error across different $f$s and different mixtures, and shading shows $[25\%,75\%]$ quantiles. The blue dot to the right represents the expectation calculated by a mean-field VI approximation (a diagonal Gaussian) using the automatic differentiation variational inference (ADVI) package built in to Stan \citep{Kucukelbir2017}, run with its default parameters.
    }
    \label{fig:posteriordb}
\end{figure*}

\section{Numerical Details}
\label{app:numerical_details}

We implemented (\ref{eqn:log_psi_stams}) in Stan \citep{Carpenter2017}. For $\q$, we used the family of multivariate Gaussians with diagonal covariance, parameterized as $\theta=[\mu_1, \ldots, \mu_n, \log\sigma_1, \ldots, \log\sigma_n]$ where $n$ is the number of unconstrained parameters (i.e the dimensionality of $\x$). In this parameterization, $\frac{1}{2}\log\FIM(\theta)$ is simply $-\sum_{i=1}^n\log\sigma_i$. We sampled $\theta$ from $\psi(\theta)$ using Stan's default implementation of the No U-Turn Sampler (NUTS) with automatic step-size adaptation \citep{Hoffman2014}, and we set the mass equal to $\lambda$ times the identity matrix. NUTS requires both $\KL(\q||\p)$ and its gradient, which we computed using Monte Carlo samples from $\q$ and the reparameterization trick. The reparameterized samples were frozen for each trajectory of NUTS and resampled between trajectories.

All code to generate the figures in this paper is available publicly online; the repository URL will be shared after the double-blind review process is complete. Python libraries used include NumPy, SciPy, PyTorch, and Matplotlib \citep{numpy,scipy,pytorch,matplotlib}.

\subsection{Figure details}

We used two toy distributions in our results:
\begin{itemize}
    \item The ``banana'' distribution over $\mathbb{R}^2$, defined as
    \begin{align*}
        \log \p(x,y) = -(y-(x/2)^2)^2 - (x/2)^2 \, .
    \end{align*}
    \item The ``Laplace mixture'' distribution over $\mathbb{R}^1$, defined as
    \begin{align*}
        \p(x) \propto 0.4e^{\frac{|x+1.5|}{0.75}} + 0.6e^{\frac{|x-1.5|}{0.75}} \, .
    \end{align*}
\end{itemize}

We also tested our method on three reference problems taken from posteriordb \citep{posteriordb}, a database of reference problems for testing and validating inference methods. These were \texttt{arK}, \texttt{eigh schools centered}, and \texttt{garch11}. Results for these additional problems are shown in Figure \ref{fig:posteriordb}.  

In our experiments, all functions integrated are sums of sinusoids, $$f(\x)=\sum_{\omega=1}^N a \sin(\omega \randomt^T\x+\phi_\omega)$$ where $\randomt$ is a random unit vector. This is a convenient target distribution as the integral of a sinusoid under a Gaussian is known analytically:
$$
\int_\x \sin(\omega \randomt^\top\x+\phi_\omega) \mathcal{N}(\mu, \Sigma) = \sin(\omega \randomt^\top\mu+\phi_\omega)\exp\left(-\frac{\omega^2}2 \randomt^T\Sigma \randomt\right)
$$
The capability for exact integration of $\int_\x\m_T(\x)f(\x)\rd\x$ ensures that no additional variance is introduced in plots; all variance is due to the selection of the components $\q$. In general this integral can be computed with MC methods or, in low enough dimensions, Gaussian quadrature.

In our experiments (Figures \ref{fig:bias_variance} and \ref{fig:wiggliness}) we used $N=100$ sinusoidal components in $f(\x)$, and calculated bias using $T=5,000$ components thinned from 4 MCMC chains of length $50,000$.
To calculate variance, we subsampled $T=10$ components from these chains, and computed variance over these random instantiations of $\m_{10}(\x)$.
The NUTS samples over $\x$ treated as ground truth derive from 4 chains of length 1,000,000.

\vfill
\pagebreak

\end{document}